\documentclass{article} 

\usepackage{iclr2024_conference,times}
\iclrfinalcopy
\usepackage{amsmath, amsthm, amssymb, amsfonts}
\usepackage{xparse}
\usepackage{bm}

\usepackage[hidelinks]{hyperref}

\usepackage{appendix}

\usepackage{soul}

\usepackage{cleveref}
\usepackage{graphicx}

\renewcommand{\top}{{\!\mathsf{T}}}

\newtheorem{theorem}{Theorem}
\newtheorem{lemma}{Lemma}

\renewcommand{\paragraph}[1]{\textbf{#1}\hspace{.5em}}

\DeclareRobustCommand{\mb}[1]{\boldsymbol{#1}}

\DeclareMathOperator*{\Tr}{Tr}

\newcommand{\mbv}{\mb{v}}

\newcommand{\mbz}{\mb{z}}

\newcommand{\mbA}{\mb{A}}

\newcommand{\mbI}{\mb{I}}

\newcommand{\mbM}{\mb{M}}

\newcommand{\mbQ}{\mb{Q}}

\newcommand{\mbS}{\mb{S}}

\newcommand{\mbX}{\mb{X}}

\newcommand{\mbgamma}{\mb{\gamma}}

\newcommand{\mbSigma}{\mb{\Sigma}}

\newcommand{\Var}{\mathbb{V}\textrm{ar}}
\newcommand{\Cov}{\mathbb{C}\textrm{ov}}

\newcommand{\bbE}{\mathbb{E}}

\newcommand{\bbP}{\mathbb{P}}

\newcommand{\bbR}{\mathbb{R}}

\newcommand{\cN}{\mathcal{N}}
\newcommand{\cO}{\mathcal{O}}

\newcommand{\cZ}{\mathcal{Z}}

\newcommand{\trans}{\mathsf{T}}

\newcommand{\reals}{\mathbb{R}}
\newcommand{\expect}{\mathbb{E}}

\renewcommand{\arccos}{\cos^{-1}}

\NewDocumentCommand{\Vc}{O{} m !g t' t"}{%
  \bm{#1{\mathbf{\MakeLowercase{#2}}}}%
  \IfValueT{#3}{_{#3}}%
  \IfBooleanTF{#4}{^{\trans}}{%
    \IfBooleanT{#5}{^{\trans*}}}%
}

\NewDocumentCommand{\Rvec}{O{} m !g t' t"}{%
  \mathslb{#1\MakeLowercase{#2}}%
  \IfValueT{#3}{_{#3}}%
  \IfBooleanTF{#4}{^{\trans}}{%
    \IfBooleanT{#5}{^{\trans*}}}%
}

\NewDocumentCommand{\Mx}{O{} m !g t' t"}{
  \bm{#1{\mathbf{\MakeUppercase{#2}}}}%
  \IfValueT{#3}{_{#3}}%
  \IfBooleanTF{#4}{^{\trans}}{%
    \IfBooleanT{#5}{^{\trans*}}}%
}

\NewDocumentCommand{\Rmat}{O{} m !g t' t"}{
  \mathslb{#1\MakeUppercase{#2}}%
  \IfValueT{#3}{_{#3}}%
  \IfBooleanTF{#4}{^{\trans}}{%
    \IfBooleanT{#5}{^{\trans*}}}%
}

\title{Estimating shape distances on neural representations with limited samples}

\author{Dean A. Pospisil$^{1}$ \: Brett W. Larsen$^{2,3,4}$ \: Sarah E. Harvey$^{2,3}$ \: Alex H. Williams$^{2,3}$ \\
$^1$Princeton University, Princeton, NJ, 08544; \: \texttt{dp4846@princeton.edu}  \\
$^2$New York University, Center for Neural Science, New York, NY, 10003 \\
$^3$Flatiron Institute, Center for Computational Neuroscience, New York, NY, 10010 \\
$^4$Flatiron Institute, Center for Computational Mathematics, New York, NY, 10010 \\ 
\texttt{\{brettlarsen, sharvey, awilliams\}@flatironinstitute.org}
}

\iclrfinalcopy 
\begin{document}

\maketitle

\begin{abstract}
Measuring geometric similarity between high-dimensional network representations is a topic of longstanding interest to neuroscience and deep learning.
Although many methods have been proposed, only a few works have rigorously analyzed their statistical efficiency or quantified estimator uncertainty in data-limited regimes.
Here, we derive upper and lower bounds on the worst-case convergence of standard estimators of \textit{shape distance}---a measure of representational dissimilarity proposed by \citet{Williams2021_shape_metrics}.
These bounds reveal the challenging nature of the problem in high-dimensional feature spaces.
To overcome these challenges, we introduce a new method-of-moments estimator with a tunable bias-variance tradeoff.
We show that this estimator achieves substantially lower bias than standard estimators in simulation and on neural data, particularly in high-dimensional settings.
Thus, we lay the foundation for a rigorous statistical theory for high-dimensional shape analysis, and we contribute a new estimation method that is well-suited to practical scientific settings.
\end{abstract}

\section{Introduction}

Many approaches have been proposed to quantify similarity in neural network representations.
Some popular methods include canonical correlations analysis \citep{Raghu2017}, centered kernel alignment \citep{Kornblith2019}, representational similarity analysis \citep[RSA;][]{Kriegeskorte2008}, and shape metrics \citep{Williams2021_shape_metrics}.
Each of these approaches takes in a set of high-dimensional measurements---e.g., hidden layer activations or neurobiological responses---and outputs a (dis)similarity score.
Shape distances additionally satisfy the triangle inequality, thus enabling downstream algorithms for clustering and regression that leverage metric space structure.

Here, we take a closer look at the estimation of shape distance in high-dimensional, noisy, and sample-limited regimes.
While shape distances have numerous applications in the physical sciences \citep{rohlf_extensions_1990, Goodall1991, andrade_procrustes_2004, Kendall2009-zb, saito_how_2015} the use of shape metrics and other measures of neural representational similarity has introduced statistical issues that have not been adequately addressed.
Specifically, shape metrics are often applied to low-dimensional noiseless measurements (e.g., 3D digital scans of anatomy across animals; \citealt{rohlf_extensions_1990}) whereas in the study of neural networks the applications have been high-dimensional  (e.g., comparing neural activity between brain regions; \citealt{Kriegeskorte2008-su}).

We demonstrate that the noise and high dimensionality of neural representations pose a substantial challenge to estimating representational similarity in sample-limited regimes. 
Yet, with the noteworthy exception of research on RSA \citep{Cai2016,Walther2016,Schutt2023}, there is little work on quantifying accuracy on estimators of representational similarity (e.g. by developing procedures to compute confidence intervals).
This poses a serious obstacle to adoption of these methods, particularly in experimental neuroscience where there is a hard limit on the number of conditions that can be feasibly sampled~\citep{Shi2019,Williams2021_single_trial}.

To address this challenge, we first obtain high-probability upper and lower bounds on the accuracy of typical ``plug-in estimates'' of shape distance.
These bounds reveal these estimators are biased with low variance.
The overall error decays with the number of sampled conditions, $M$, but the decay rate is inversely related to the number of dimensions, $N$.
To combat these limitations, we propose a new method-of-moments estimator which, while not always strictly optimal compared with the plug-in estimator,  provides an explicit and tunable tradeoff between estimator bias and variance.

\begin{figure}
\centering
\includegraphics[width=0.8\linewidth]{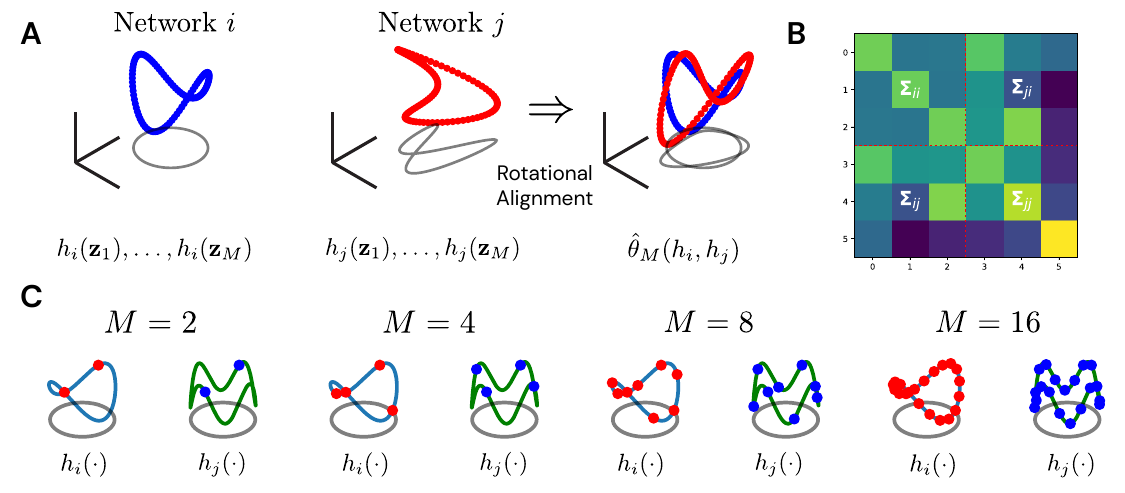}
\vspace{-0.25cm}
\caption{\textbf{(A)} Classical shape distances \citep{Kendall2009-zb} can be used to provide a rotation-invariant distance between neural representations~\citep{Williams2021_shape_metrics}.
Given two labelled points clouds in $N$-dimensional space (\textit{left} and \textit{middle}), the distance is computed after an optimal orthogonal transformation is chosen to align the point clouds (\textit{right}). In this visual example the point clouds trace out a low-dimensional manifold. \textbf{(B)} Heatmap shows the covariances ($\mbSigma_{ii}, \mbSigma_{jj}$) and cross-covariance ($\mbSigma_{ij}$) of the 3D representations in panel A.
Shape distances can be re-expressed in terms of these quantities (see eq.~\ref{eq:procrustes-alt}, \ref{eq:riemannian-alt}). \textbf{(C)} Our ability to estimate the shape distance is related to $M$, the number of stimuli.
As $M$ increases (\textit{left} to \textit{right}) the number of sampled points along the underlying manifold increases, and we are better able to resolve shape differences between the representations.} 
\vspace{-1.2em}
\label{fig:interpolate}
\end{figure}
\section{Background and Problem Setting}

We begin by considering a simple setting where each neural network is a deterministic map (for the stochastic setting, see \cref{subsec:stochastic}).
A collection of $K$ neural networks can then be viewed as a set of functions, each denoted ${h_i : \cZ \mapsto \reals^{N}}$ for $i \in \{1, \dots, K\}$.
Here, $\cZ$ is a feature space and $N$ can be interpreted as the number of neurons in each system (e.g. the size of a hidden layer in an artificial network, or the number of recorded neurons in a biological experiment).\footnote{The assumption that each layer has the same number of neurons is not essential. A theoretical connection with the Bures distance pointed out by \citet{harvey_duality_2023} allows one to generalize shape distances to networks of dissimilar sizes. Indeed, we will see in \cref{eq:procrustes-alt,eq:riemannian-alt} how shape distances can be expressed in terms of covariance and cross-covariance matrices that are well-defined in unequal dimensions.}

Let $h_i$ and $h_j$ denote neural systems which we assume are mean-centered and bounded:
\begin{equation}
\label{eq:constraints}
\expect [h_i(\mbz)] = \expect [h_j(\mbz)] = \mb{0} \hspace{2em} \text{and} \hspace{2em} \Vert h_i(\mbz) \Vert_2, \Vert h_j(\mbz) \Vert_2 < B \sqrt{N} \quad \text{almost surely.}
\end{equation}
for some constant $B > 0$.
Here, the expectations are taken over $\mbz \sim P$, for some distribution $P$ over network inputs.
Our assumption that neural population rates are bounded by $B\sqrt{N}$ can result from assuming each neuron has a maximum firing rate equal to $B$.
This assumption is common in the literature and reasonable in both artificial networks (since connection weights are finite) and biological networks (since neurons have a maximal firing rate).

Motivated by the shape theory literature~\citep{Goodall1991, Kent1997, Kendall2009-zb,Williams2021_shape_metrics}, we consider estimating the \textit{Procrustes size-and-shape distance}, $\rho$, and \textit{Riemannian shape distance}, $\theta$, between neural representations.
In our setting, these shape distances can be defined as \citep[see App. D in][]{Williams2021_shape_metrics}:
\begin{align}
\label{eq:procrustes}
\rho(h_i, h_j) &= \min_{\mbQ \in \cO(N)} \sqrt{\expect \Vert h_i(\mbz) - \mbQ h_j(\mbz) \Vert_2^2} \\
\label{eq:riemannian}
\theta(h_i, h_j) &= \min_{\mbQ \in \cO(N)} \arccos \left ( \frac{ \expect [ h_i(\mbz)^\top \mbQ h_j(\mbz) ]}{\sqrt{\expect[h_i(\mbz)^\top h_i(\mbz)] \expect [h_j(\mbz)^\top h_j(\mbz)]}} \right )
\end{align}
where $\cO(N)$ denotes the set of $N \times N$ orthogonal matrices.
Again, all expectations are taken over $\mbz \sim P$.
Note that different notions of distance arise from different choices of input distribution, $P$.

To simplify our analysis and exposition, we will focus on estimating the \textit{squared Procrustes distance}, $\rho^2$, and what we call the \textit{cosine shape similarity}, $\cos \theta$.
Thus, we ignore the square root term in \cref{eq:procrustes} and the arccosine term in \cref{eq:riemannian}, but it should be kept in mind that one must apply these nonlinear functions to achieve a proper metric.

\paragraph{Properties of Shape Distance}
It is easy to verify that shape distances are invariant to rotations and reflections: that is, if ${r : \reals^N \mapsto \reals^N}$ is an orthogonal transformation, then for any function ${h : \cZ \mapsto \reals^N}$ representing a neural system we have ${\rho(h, r \circ h) = \theta(h, r \circ h) = 0}$, where `$\circ$' denotes function composition.
Furthermore, $\rho$ and $\theta$ are proper metrics, meaning that:
\begin{equation}
\rho(h_i, h_j) = \rho(h_j, h_i) \hspace{1em} \text{and} \hspace{1em} \rho(h_i, h_j) \leq \rho(h_i, h_k) + \rho(h_k, h_j) \hspace{1em} \forall~ i, j, k \in \{1, \dots, K\},
\end{equation}
and likewise for $\theta$.
These properties are fundamental to rigorously establishing downstream analyses, such as for clustering networks with similar representations~\citep{Williams2021_shape_metrics}.

It is well-known that the optimal orthogonal alignment appearing in \cref{eq:procrustes,eq:riemannian} can be identified in closed form, allowing us to write the Procrustes and Riemannian shape distances in terms of the covariance and cross-covariance matrices.
We define the covariance ($\Sigma_{ii}$ and $\Sigma_{jj}$) and cross-covariance matrices ($\Sigma_{ij}$)  as 
\begin{equation}
\mbSigma_{ii} = \expect [ h_i(\mbz) h_i(\mbz)^\top ]
~~ , ~~
\mbSigma_{jj} = \expect [ h_j(\mbz) h_j(\mbz)^\top ]
~~ , ~~
\mbSigma_{ij} = \expect [ h_i(\mbz) h_j(\mbz)^\top ],
\end{equation}
and reformulate the squared Procrustes distance and cosine shape similarity:
\begin{align}
\label{eq:procrustes-alt}
\rho^2(h_i, h_j) &= \Tr[\mbSigma_{ii}] + \Tr[\mbSigma_{jj}] - 2 \Vert \mbSigma_{ij} \Vert_* \\
\label{eq:riemannian-alt}
\cos \theta(h_i, h_j) &= \frac{ \Vert \mbSigma_{ij} \Vert_* }{\sqrt{\Tr[\mbSigma_{ii}] \Tr[\mbSigma_{jj}]}}
\end{align}
where $\Vert \mbSigma_{ij} \Vert_*$ denotes the nuclear norm (or Shatten 1-norm) of the cross-covariance matrix:
\begin{equation}
\Vert \mbSigma_{ij} \Vert_* = \sum_{n=1}^N s_n(\mbSigma_{ij})
\end{equation}
where $s_1(\mbM) \geq \dots \geq s_N(\mbM) \geq 0$ denote the singular values of a matrix $\mbM$.
Equations~\ref{eq:procrustes-alt} and \ref{eq:riemannian-alt} are derived in Appendix~\ref{app:reformulate} to provide the reader with a self-contained narrative.

\paragraph{Plug-in Estimators}
Suppose we are given $M$ independent and identically distributed network inputs $\mbz_1, \dots, \mbz_M \sim P$.
How well can we approximate the shape distances between two networks, as a function of $M$?
The standard approach \citep{Williams2021_shape_metrics}, is to use a \textit{plug-in estimator} in which one computes \cref{eq:procrustes,eq:riemannian} after identifying the optimal $\mbQ \in \cO(N)$.
As we show in App.~\ref{app:convert-plug-in}, this is equivalent to estimating the squared Procrustes and cosine Riemannian distances by substituting the empirical covariances:
\begin{equation}
\hat{\mbSigma}_{ii} = {\textstyle \frac{1}{M} \!\! \sum\limits_{m=1}^M} h_i(\mbz_m) h_i(\mbz_m)^\top
,~
\hat{\mbSigma}_{jj} = {\textstyle \frac{1}{M} \!\! \sum\limits_{m=1}^M} h_j(\mbz_m) h_j(\mbz_m)^\top
,~
\hat{\mbSigma}_{ij} = {\textstyle \frac{1}{M} \!\! \sum\limits_{m=1}^M} h_i(\mbz_m) h_j(\mbz_m)^\top 
\end{equation}
to approximate the true covariances appearing in \cref{eq:procrustes-alt,eq:riemannian-alt}.
Thus,
\begin{align}
\label{eq:procrustes-plug-in-estimator}
\hat{\rho}^2(h_i, h_j) &= \Tr[\hat{\mbSigma}_{ii}] + \Tr[\hat{\mbSigma}_{jj}] - 2 \Vert \hat{\mbSigma}_{ij} \Vert_* \\
\label{eq:riemannian-plug-in-estimator}
\cos \hat{\theta}(h_i, h_j) &= \frac{ \Vert \hat{\mbSigma}_{ij} \Vert_* }{\sqrt{\Tr[\hat{\mbSigma}_{ii}] \Tr[\hat{\mbSigma}_{jj}]}}
\end{align}
define plug-in estimators for the squared Procrustes and cosine Riemannian shape distances.
The empirical behavior of these estimators as a function of $M$ was only briefly characterized by \citet{Williams2021_shape_metrics} for a pair of artificial networks trained on CIFAR-10.

\section{Results}

First, we theoretically characterize the accuracy of plug-in estimation as a function of the number of samples, $M$, and the dimension, $N$.
We show that these estimators are biased and can converge at unfavorably slow rates under certain conditions.
To overcome these issues, we introduce a new method-of-moments estimator in \cref{subsec:method-of-moments-estimator} which has lower bias at the cost of increased variance.

\subsection{Nonasymptotic bounds on the performance of plug-in estimation}
\label{subsec:upper-bound-on-plugin}

First, it is straightforward to estimate $\Tr [\mbSigma_{ii}]$ and $\Tr [\mbSigma_{jj}]$.
Their plug-in estimators are unbiased under our assumptions in \cref{eq:constraints}, and they rapidly converge to the correct answer. This is shown in the following lemma, whose proof relies only on classical concentration inequalities.
\begin{lemma}[App.~\ref{app:plug-in-covaraince}]
\label{lemma:tail-bound-for-covariance}
Under the assumptions in \cref{eq:constraints}, with probability at least $1-\delta$:
\begin{equation}
    \left | \Tr[\mbSigma_{ii}] - \Tr [ \hat{\mbSigma}_{ii} ] \right | \leq B N^{1/2}M^{-1/2} \sqrt{2\log(2/\delta)}
\end{equation}
\end{lemma}
In contrast, the plug-in estimator for $\Vert \mbSigma_{ij} \Vert_*$ is biased upwards (see \cref{subsec:lower-bound-results}) and turns out to converge more slowly.
Using the Matrix Bernstein inequality~\citep[see][]{Tropp2015}, we can show:
\begin{lemma}[App.~\ref{app:upper-cross-covariance}]
\label{lemma:upper-bound-nuc-norm}
Under the assumptions in \cref{eq:constraints}, for any $M$ and $N$:
\begin{equation}
\label{eq:lemma-upper-bound-nuc-norm}
\expect  \Big | \Vert \hat{\mbSigma}_{ij} \Vert_* - \Vert \mbSigma_{ij} \Vert_* \Big | < \frac{2B^2 N^2 \log(2N)}{3M} + \frac{2 B^2 N^2 \sqrt{\log(2N)}}{M^{1/2}}
\end{equation}
\end{lemma}
This only upper bounds the expected error.
However, the fluctuations around this expectation turn out to be small (see App.~\ref{app:tail-cross-covariance}), and so we are able to combine lemmas~\ref{lemma:tail-bound-for-covariance} and~\ref{lemma:upper-bound-nuc-norm} into the following:
\begin{theorem}[App.~\ref{app:tail-cross-covariance}]
\label{theorem:upper-bound-on-squared-procrustes}
Under the assumptions in \cref{eq:constraints},
with probability at least $1 - \delta$
\begin{equation}
\frac{|\hat{\rho}^2 - \rho^2|}{N} \leq \frac{2 B^2 N \log(2N)}{3M} + \frac{2 B^2 N \sqrt{\log(2N)}}{M^{1/2}} + \left ( \frac{B^2}{M^{1/2}} + \frac{2B}{N^{1/2}M^{1/2}} \right ) \! \! \sqrt{2 \log \left ( \frac{6}{\delta} \right )}
\end{equation}
\end{theorem}
\Cref{theorem:upper-bound-on-squared-procrustes} states a non-asymptotic upper bound on the plug-in estimator's error that holds with high probability.
We have expressed this bound on the squared size-and-shape Procrustes distance normalized by $1/N$, since the raw error, $|\hat{\rho} - \rho|$, will tend to increase linearly with $N$ for an uninteresting reason---namely, since the the Procrustes shape distance is comprised of terms like $\Tr[\mbSigma_{ii}]$ and $\Tr[\mbSigma_{jj}]$.
The choice of normalization in \cref{theorem:upper-bound-on-squared-procrustes} also makes the result more comparable to the cosine shape similiarity (eq.~\ref{eq:riemannian-alt}), which is normalized by a factor, $\sqrt{\Tr[\mbSigma_{ii}] \Tr[\mbSigma_{jj}]}$, of order $N$.

We can gain intuition for \cref{theorem:upper-bound-on-squared-procrustes} by ignoring logarithmic factors and noticing that the second term dominates.
Then, roughly speaking, \cref{theorem:upper-bound-on-squared-procrustes} says that we can guarantee the plug-in error decreases as a function of $N M^{-1/2}$.
Thus, for any fixed $N$, we need to increase $M$ by a factor of 4 to decrease estimation error by a factor of 2.
Further, when comparing higher-dimensional neural representations (i.e. higher $N$) we need to sample more landmarks---if $N$ increases by a factor of 2, then $M$ must be increased by a factor of 4 to compensate.

\subsection{Failure modes of plug-in estimation and a lower bound on performance}
\label{subsec:lower-bound-results}

\Cref{theorem:upper-bound-on-squared-procrustes} provides a high probability upper bound on the estimation error.
A natural question is whether this upper bound is tight.
To investigate, we seek an example where the plug-in estimator performs badly.
We intuited that the plug-in estimates will have a large downward bias when two neural representations are very far apart in shape space.
This can be understood in two ways.
First, from the definitions of $\rho$ and $\theta$ in \cref{eq:procrustes,eq:riemannian}, we see that both expressions contain a minimization over $\mbQ \in \cO(N)$.
For large $N$ and small $M$, this high-dimensional orthogonal matrix can be ``overfit'' to the $M$ observations resulting in an underestimate of distance.
Second, from the alternative formulations in \cref{eq:procrustes-alt,eq:riemannian-alt}, we see that the shape distance is large if the true cross-covariance is ``small'' as quantified by the nuclear norm.
In the extreme case where the singular values of $\mbSigma_{ij}$ are all zero, the empirical cross-covariance matrix $(1/M)\sum_m h_i(\mbz_m) h_j(\mbz_m)^\top$ will overestimate the nuclear norm, and therefore underestimate the shape distance.
This is more severe when $M$ is small, since there are fewer terms in the sum to ``average out'' spurious correlations, which are particularly problematic in high dimensions (i.e. when $N$ is large).

This intuition led us to construct an example where plug-in estimation error approaches the upper bound in \cref{theorem:upper-bound-on-squared-procrustes}.
This is summarized in the following result.

\begin{theorem}[Lower Bound, App.~\ref{app:lower-cross-covaraince}]
\label{theorem:lower-performance-bound}
Under the assumptions in \cref{eq:constraints}, there exist neural networks and a distribution over inputs such that in the limit that $N \rightarrow \infty$ and $M \gg N$:
\begin{equation}
\label{eq:theorem-lower-performance-bound}
\frac{|\hat{\rho}^2 - \rho^2|}{N} = \frac{16 B^2}{3 \pi} N^{1/2} M^{-1/2}
\end{equation}
\end{theorem}
In Appendix~\ref{appendix:numerical-verification-of-lower-bound} we show the validity of this lower bound on simulated data.
Although the bound is asymptotic, it gives a highly accurate approximation to the observed plug-in error for reasonable values of $M$ and $N$ (see Fig.~\ref{fig:appendix-lower-bound-numerical-verification}).
Thus, while future work may seek to improve the upper bound in \cref{theorem:upper-bound-on-squared-procrustes}, we cannot hope to improve beyond the lower bound formulated above.
If we ignore constant factors and logarithmic terms to gain intuition, we observe there is (roughly) a gap of $N^{1/2}$ between the upper and lower bounds.
Thus, it is possible that our analysis in \cref{subsec:upper-bound-on-plugin} may be conservative in terms of the ambient dimension.
That is, to compensate for a two-fold increase in $N$, \cref{theorem:lower-performance-bound} only shows a case where $M$ needs to be increased two-fold, in contrast to the four-fold increase suggested by \cref{theorem:upper-bound-on-squared-procrustes}.
However, in terms of the number of sampled inputs, the lower and upper bounds match: thus, the rate cannot be improved beyond $M^{-1/2}$.

\subsection{A new estimator with controllable bias}
\label{subsec:method-of-moments-estimator}

The plug-in estimator of $\Vert \mbSigma_{ij} \Vert_*$ has low variance but large and slowly decaying bias (see theorems \ref{theorem:upper-bound-on-squared-procrustes} and \ref{theorem:lower-performance-bound}).
Here we develop an alternative estimator that is nearly unbiased.

First, note that the eigenvalues of $\mbSigma_{ij} \mbSigma_{ij}^\top$ correspond to the squared singular values of $\mbSigma_{ij}$.
Thus, $\Tr[(\mbSigma_{ij} \mbSigma_{ij}^\top)^{1/2}] = \Vert \mbSigma_{ij} \Vert_*$, and so we can reduce our problem to estimating the trace of $(\mbSigma_{ij} \mbSigma_{ij}^\top)^{1/2}$, which is symmetric.
Leveraging ideas from a well-developed literature~\citep{Adams2018}, we proceed to define the $p^\text{th}$ moment of this matrix as:
\begin{equation}
W_p = \Tr [ (\mbSigma_{ij} \mbSigma_{ij}^\top)^p ] = \sum_{n=1}^N \lambda_n^p
\end{equation}
where $\lambda_1, \dots, \lambda_N$ denote the eigenvalues of $\mbSigma_{ij} \mbSigma_{ij}^\top$.
Now, for any function $f : \reals \mapsto \reals$ and symmetric matrix $\mbS$ with eigenvalues $\lambda_1, \dots, \lambda_N$, we define\footnote{This is a common convention to extend scalar functions \citep[see e.g.][sec. 1.2.6]{potters_bouchaud_2020}.} $\Tr[f(\mbS)] = \sum_{i} f(\lambda_i)$.
So long as $f$ is reasonably well-behaved, we can approximate it using a truncated power series with $P$ terms.
Thus, with $\mbS = \mbSigma_{ij} \mbSigma_{ij}^\top$ and $f(x) = \sqrt{x}$:
\begin{equation}
\label{eq:power_series_approximation}
\Vert \mbSigma_{ij} \Vert_* = \Tr[(\mbSigma_{ij} \mbSigma_{ij}^\top)^{1/2}] \approx \sum_{n=1}^N \sum_{p=0}^P \gamma_p \lambda_n^p = \sum_{p=0}^P \gamma_p \sum_{n=1}^N \lambda_n^p =  
\sum_{p=0}^P \gamma_p W_p
\end{equation}
where $\gamma_0, \dots, \gamma_P$ are scalar coefficients.

In summary, we can estimate $\Vert \mbSigma_{ij} \Vert_*$ by (a) specifying an estimator of the top eigenmoments, $W_1, \dots, W_P$, and (b) specifying a desired set of scalar coefficients $\gamma_0, \dots, \gamma_P$.
To estimate the eigenmoments, we adapt procedures described by \citet{Kong2017} to obtain unbiased estimates for each moment, $\hat{W}_1, \dots, \hat{W}_P$ (see App.~\ref{app:wp-ubiased}).
To select the scalar coefficients, we propose an optimization procedure that trades off between bias and variance in the estimate of $\Vert \mbSigma_{ij} \Vert_*$.
Our starting point is the usual bias-variance decomposition:
\begin{equation}
\expect \left [ \left ( \Vert \mbSigma_{ij} \Vert_* - \textstyle\sum_p \gamma_p \hat{W}_p \right )^2 \right ] = \left( \expect \left [ \Vert \mbSigma_{ij} \Vert_* - \textstyle\sum_p \gamma_p \hat{W}_p \right ] \right )^2 + \Var \left [\textstyle\sum_p \gamma_p \hat{W}_p \right ].
\end{equation}
Since $\expect [\hat{W}_p] = W_p = \sum_n \lambda_n^p$, the first term above (i.e. the ``bias'') simplifies and is upper-bounded:
\begin{equation*}
\left( \expect \left [ \Vert \mbSigma_{ij} \Vert_* - \textstyle\sum_p \gamma_p \hat{W}_p \right ] \right )^2 = \left( \textstyle\sum_n \left ( \lambda_n^{1/2} - \textstyle\sum_p \gamma_p \lambda_n^p \right ) \right )^2 \leq \max_{0 \leq x \leq 1} \left ( N \left ( x^{1/2} - \textstyle\sum_p \gamma_p x^p \right ) \right )^2
\end{equation*}
The inequality follows from replacing each term in the sum over $n$ with the worst case approximation error of the polynomial expansion (given here as the maximization over $x$).
Thus, we seek to:
\begin{equation}
\label{eq:gamma-quad-prog}
\underset{\gamma_0, \dots, \gamma_P}{\text{minimize}} \quad  \max_{0 \leq x \leq 1} \left ( N  \left ( x^{1/2} - \textstyle\sum_p \gamma_p x^p \right ) \right)^2 +  \textstyle\sum_{p, p^\prime} \gamma_p \gamma_{p^\prime} \Cov (\hat{W}_p, \hat{W}_{p^\prime}).
\end{equation}
We estimate $\Cov (\hat{W}_p, \hat{W}_{p^\prime})$ by bootstrapping---i.e. the empirical covariance of these statistics across re-sampled datasets where $\{\mbz_1, \dots, \mbz_M\}$ are sampled with replacement.
Given this estimate of covariance, \cref{eq:gamma-quad-prog} can be cast as a convex quadratic program and the maximal bias can be bounded to a user defined limit at the expense of variance (see App.~\ref{app:quadratic-program}). 
 We use the maximal bias (eq. \ref{eq:gamma-quad-prog}, term 1) and variance (eq. \ref{eq:gamma-quad-prog}, term 2) to form approximate confidence intervals (see App.~\ref{app:confidence-intervals}). 
\subsection{Extension to stochastic networks}
\label{subsec:stochastic}

Thus far, we have modeled neural networks as deterministic mappings, $h_i : \cZ \mapsto \reals^N$.
This assumption is not satisfied in biological data and in many artificial networks (e.g. VAEs).
Here, we briefly explain how to extend the estimators to the stochastic setting.
In this setting, the response of network $i$ can be written as ${h_i(\mbz) + \epsilon_i(\mbz)}$.
As before, $h_i(\mbz)$ is a deterministic mapping conditioned on a random variable $\mbz \sim P$.
The ``noise'' term $\epsilon_i(\mbz)$ is a mean-zero random variable that, in addition to inheriting the randomness of $\mbz$, captures the stochastic elements of each forward pass through the network (i.e. trial-to-trial variability even when the stimulus is fixed).
Importantly, noise contributions are independent and identically distributed for each pass through the network.

Given a second stochastic network with same structure, ${h_j(\mbz) + \epsilon_j(\mbz)}$, our goal is to estimate the shape distances~\cref{eq:procrustes,eq:riemannian} as before, effectively ignoring contributions of the ``noise'' terms $\epsilon_i(\cdot)$ and $\epsilon_j(\cdot)$.
Ignoring these terms is not wholly justified, since it is of great interest to quantify how noise varies across networks \citep{Duong2023}.
Nonetheless, it is useful to develop metrics that isolate the ``signal'' component of neural representations, and a full development of methods to quantify similarity in noise structure is outside the scope of this paper.

Our basic observation is that it suffices to consider two replicates for each network input.
That is, let $\mbz^\prime = \mbz$ where $\mbz \sim P$.
Then, $\mbSigma_{ii} = \expect[h_i(\mbz) h_i(\mbz^\prime)^\top]$ which can be approximated by the slightly reformulated plug-in estimator: $\mb{\hat\Sigma}_{ii} = (1/M) \sum_m h_i(\mbz_m) h_i(\mbz_m^\prime)^\top$.
Further, since noise is independent across networks, i.e. $\epsilon_i(\mbz) \perp \!\!\! \perp \epsilon_j(\mbz)$ for all $\mbz \in \cZ$, the cross-covariance estimators, including the method-of-moments estimator described in \cref{subsec:method-of-moments-estimator}, do not require any modification.

\section{Applications and Experiments}
\label{sec:applications-and-experiments}

\subsection{Validation on synthetic data}
\label{subsec:simulated-data-application}

We first validate our method-of-moments estimator (\cref{subsec:method-of-moments-estimator}) on simulated responses from a multivariate normal distribution. We estimate the cosine shape similarity, $\cos \theta$, defined in eq.~\ref{eq:riemannian-alt}. Our estimator of $ \Vert \mbSigma_{ij} \Vert_*$ is the principle novelty; thus, it is informative to understand its properties in isolation. To achieve this, we use the ground truth covariance of $\hat{W}_p$ (instead of an estimate from a bootstrap) and use the ground truth values of $\Tr[\mbSigma_{ii}]$ and $\Tr[\mbSigma_{jj}]$. For details see App. \ref{app:simulations}.

\begin{figure}[htb]
\centering
\includegraphics[width=.9\linewidth]{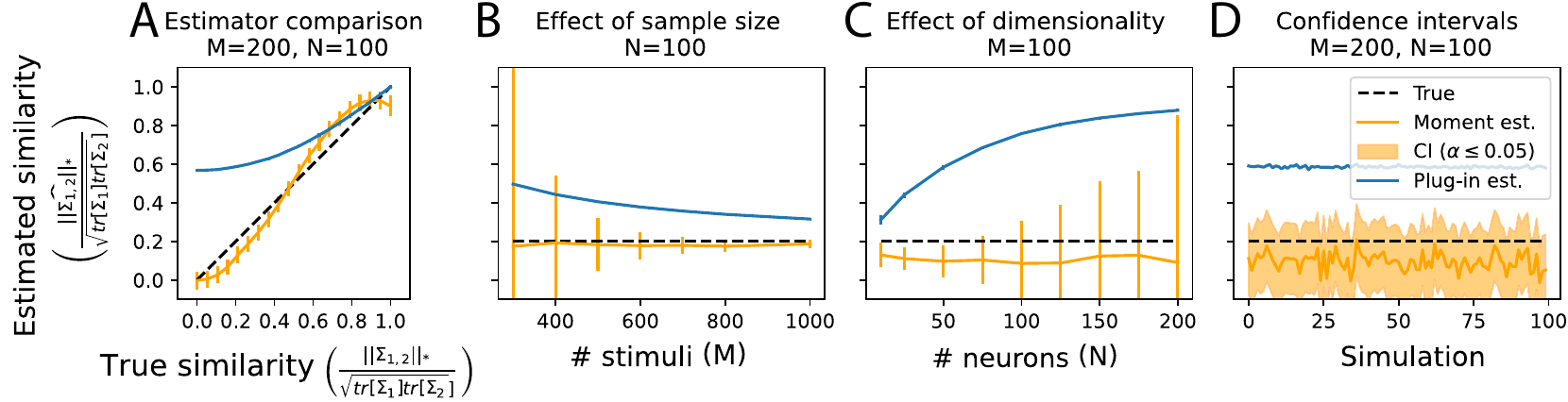}
\vspace{-0.4cm}
\caption{Validation of estimator on synthetic data. \textbf{(A)} The moment based estimator (orange) compared to plug-in estimator (blue) in simulation with standard deviation bars calculated across simulations. Estimators are evaluated at 20 linearly spaced ground truth similarity score values. \textbf{(B)} Effect of increasing sample size when moment estimator is constrained to have a bias less than 5 \%. \textbf{(C)} Effect of increasing dimensionality. \textbf{(D)}  Demonstration of conservative confidence intervals that account for variance and maximal bias of moment estimator. We do not include CIs for the plug in estimator (implied by \cref{theorem:upper-bound-on-squared-procrustes}) because for small sample sizes, the theoretical bounds on estimator bias always contain far more than the entire allowable interval ($[0,1]$). }
\vspace{-1em}
\label{fig:synthetic}
\end{figure}

We first compared the bias of the plug-in estimator to that of the moment-based estimator across a range of ground truth shape similarity values (Fig.~\ref{fig:synthetic}A). As expected from our intuition discussed in \cref{subsec:upper-bound-on-plugin}, the plug-in estimator (blue line) tends to  inflate estimated similarity when ground truth is low (left side of plot).
The moment-based estimator (orange line), in contrast, performs well over the full range, at the cost of increases in estimator variance (blue vs orange error bars).

Next, we fixed the ground truth similarity at 0.2 and studied the effect of sample size, $M$ (Fig.~\ref{fig:synthetic}B). The moment estimator (constrained to 5\% bias) maintains small bias even with small $M$, at the cost of high variance (orange error bars). Increasing $M$ quickly reduces the variance of the estimator. A similar story emerges when we fix $M$ and vary the dimension $N$ (Fig.~\ref{fig:synthetic}C).
As the dimensionality increases, the plug-in estimator bias quickly explodes. In contrast, the moment estimator (constrained to 10\% bias) has roughly constant bias; but it's variance grows with $N$. Thus our estimator bias outperforms the plug-in when the sample size is low and dimensionality is high.

Finally, an important property of the moment-based estimator is our ability to compute approximate confidence intervals (CI) (see App. \ref{app:confidence-intervals}).
We demonstrate 95\% CIs across simulations in Figure~\ref{fig:synthetic}D. These CIs are conservative, the true shape score is not within the CI's for only 2.3\% of simulations. 

\begin{figure}[ht]
\centering
\includegraphics[width=.9\linewidth]{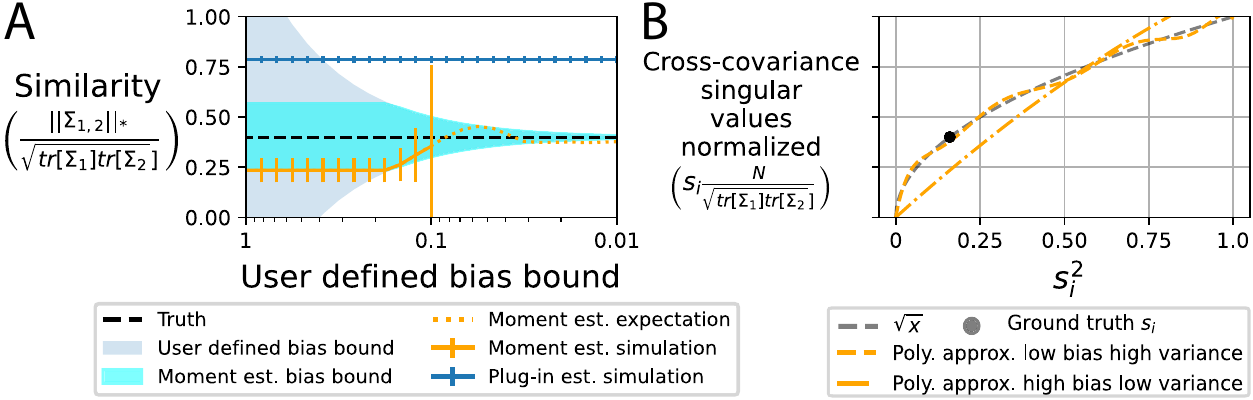}
\vspace{-0.1cm}
\caption{Control of bias-variance tradeoff with user defined bound on bias. \textbf{(A)} Moment based estimator expected value is constrained to be within the user defined bias bound (blue region) while minimizing worst case MSE (\cref{eq:gamma-quad-prog}). Maximal bias can be less than the user defined bias (cyan region within blue). Lower bias leads to increased variance (orange trace converges to black dashed as SD bars widen). Where simulations become unstable we plot the theoretical expected value (dotted orange). Plug-in estimator is well outside bias bounds of moment estimator thus is more biased than moment estimator (blue trace outside cyan line). \textbf{(B)} Example plots of solutions to the quadratic program's approximation (orange traces) to square root (black dashed trace) of the eigenvalues of $\Sigma_{1,2}$ (black points). Re-scaling of singular values on vertical axis results in the deviation between the polynomial and the true square root evaluated at the true eigenvalues being exactly the bias of the associated estimates in panel A. }
\label{fig:bias_control}
\vspace{-1em}
\end{figure}

\paragraph{Control of estimator bias}
Here we demonstrate the bias-variance tradeoff controlled by the upper-bound on bias defined by the user.  The quadratic program in \cref{eq:gamma-quad-prog} constraings the maximal absolute bias below a chosen constant (Fig.~\ref{fig:bias_control}A, blue shaded area around true similarity score). The actual maximal bias will then be less than or equal to this user defined bias (cyan shaded area within blue). The expected value of the moment estimator stays within the maximal bias, in this case on its bound (orange trace). The user defined bias bound remains inactive until it is less than the MSE minimizing solution's bias (blue completely overlapped by cyan above ~0.1). Variance then begins to increase as higher order $W_p$ terms are weighted more to reduce bias (orange standard deviation bars from simulation increase as cyan region narrows). The mean of the estimator converges to ground truth as it is constrained by the bias bound (dotted orange line converges to dashed black). The plug-in estimator exceeds the maximal bias of the moment estimator (blue trace above cyan area). 

Intuition for the moment estimator can be drawn from plots of solutions to the polynomial approximation (eq. \ref{eq:power_series_approximation}, Fig.~\ref{fig:bias_control}B, orange trace approximates black dashed) of the squared singular values of $\Sigma_{1,2}$ (black points all overlapping). Here we have re-scaled the  the vertical axis so that the deviation between the square root and polynomial approximation is exactly the bias of the moment estimator. In the case where bias is not constrained (associated with left most estimates in panel B) the approximation is poor (dashed-dot orange trace does not match dashed black trace). For these eigenvalues the the deviation is near the worst possible bias (distance from black point dashed dot orange line is nearly as far as any other vertical deviation between the traces), this is why the estimator in panel B sits at the bound of maximal possible bias. When the upper bound on bias is very small (far right of B) the approximation is very good (dashed orange overlaps dashed black) because higher order terms are used. Yet this results in very high variance (Fig.~\ref{fig:bias_control}B).

\subsection{Application to biological data}
\label{subsec:biological-data-application}

Here we investigate noisy non-Gaussian data where the covariance of the $\hat{W}_p$ and the denominator of the similarity score must be estimated from data. We do so by applying our estimator to neural data: calcium recordings from mouse primary visual cortex in responses to a set of 2,800 natural images repeated twice \citep{stringer2019high}. Our estimator became highly variable when applied to this data in part because of its low SNR (average SNR $\approx 0.1$). We therefore restricted our analysis to the neurons with the highest SNR in each recording (80 neurons in each recording).

This dataset contains seven recordings from different animals, but the population responses are not directly comparable since each recording targets a different region of primary visual cortex containing neurons with different receptive field.
Thus, even though the same images were shown across recording sessions, the recorded neurons were effectively responding to different cropped portions of the image.
We therefore only quantified shape similarity between subsampled populations of neurons taken from the same recording session.

Determining the properties of the bias of our estimator requires comparison to the ground truth value of the similarity score. In the neural data, ground truth is unknown. We thus developed two sampling schemes to set the ground truth similarity in the neural data. To set similarity to 0 we measured similarity between different subpopulations of neurons ($N=40$  neurons each) shown different stimuli (M=400 stimuli each), thus the two populations responses are independent, thus their cross covariance is 0 so that the similarity score is 0. To set the similarity to 1 we measured similarity between the same subpopulation of neurons ($N=40$ neurons) shown the same stimuli ($M=400$) but on different trials, thus the only deviation in their responses is owing to trial-to-trial variability, thus their tuning similarity is 1. 

\begin{figure}[ht]
\centering
\includegraphics[width=.8\linewidth]{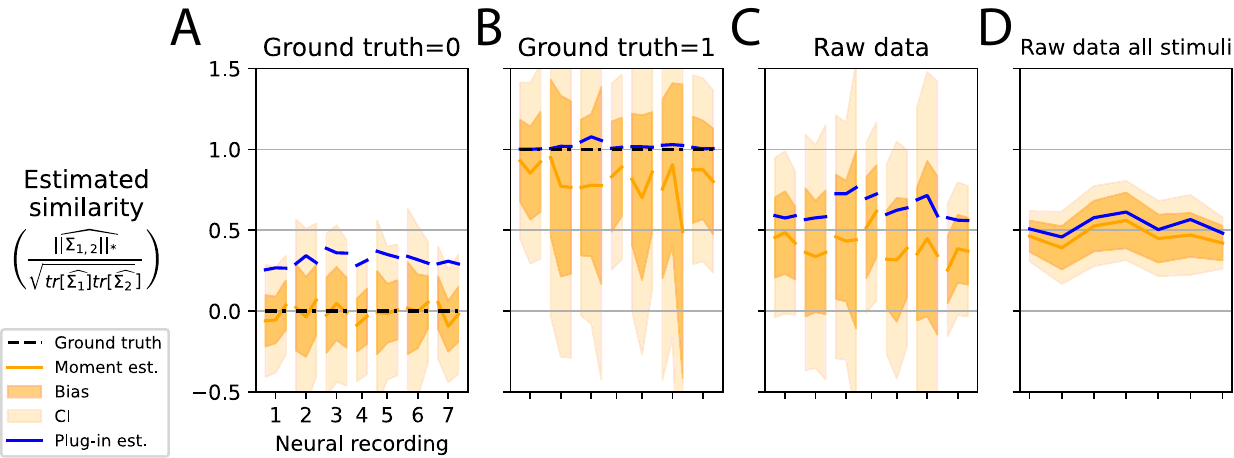}
\vspace{-0.25cm}
\caption{Validation of estimator on neural data \citep{stringer2019high}. \textbf{(A)}  Comparison of estimators when ground truth similarity of neural data is set to 0. The estimator is applied to three disjoint sets of random stimuli for each recording ($n=7$). The estimated maximal bias is plotted in dark orange area and the confidence interval, which includes bias, is plotted in light orange.  \textbf{(B)} Same simulation as (A) except ground truth similarity is 1.\textbf{(C)} Same as (B) except estimation of true similarity. \textbf{(D)} Estimation of true similarity on all stimuli. ($M\approx 2,800$).}
\label{fig:neural_data}
\vspace{-0.25cm}
\end{figure}

When the ground truth similarity was 0, the moment estimator correctly indicated this outcome (Fig.~\ref{fig:neural_data}A, orange trace overlaps black dashed) and the confidence intervals always contained the true similarity (light orange contains black dashed). On the other hand the plug-in estimator was upwardly biased (blue above black dashed). Thus the moment based estimator can accurately determine when the similarity is low in noisy neural data whereas the plug-in estimator cannot.

When ground truth similarity was 1, we found the bias of the moment estimator was worse than that of the plug-in (Fig.~\ref{fig:neural_data}B, blue overlaps black dashed, orange below). This is consistent with our synthetic simulations (see Fig.~\ref{fig:synthetic}A far right). The CIs always contained the true value but contained nearly the entire possible range of similarity values. Thus while the average estimate is high our confidence intervals are so wide that we do not have much information about the true similarity. 

Finally, we aimed to estimate the true shape similarity between these sub-populations of high SNR neurons ($N=40$ neurons each). In Figure~\ref{fig:neural_data}C, we show the estimated similarity across three independent folds of the stimulus set ($M=400$ stimuli each). Across all seven recordings the moment estimator was near 0.5, but confidence intervals were wide so there is little information about similarity even for the highest SNR neurons (light orange extends from 0 to 1 on vertical axis). The plug-in estimator reports a higher degree of similarity, that we heavily discount given its upward bias (documented in Fig.~4\ref{fig:neural_data}A). When we included all stimuli ($M\approx 2800$) we obtained tighter confidence intervals, learning that the true similarity is most likely between 0.25 and 0.75 (Fig.~\ref{fig:neural_data}D).  Thus small populations of well-tuned neurons in the same brain region have only intermediate levels of representational similarity. Overall, we find noisy data is a challenging setting for reducing the bias of shape similarity estimates. 

\subsection{Application to artificial neural network representations}

In Appendix~\ref{appendix:artificial-neural-net} we apply the plug-in and moment-based estimator to penultimate layer representations between two ResNet-50 architectures~\citep{he2016deep} trained on ImageNet classification~\cite{deng2009imagenet}.
In this setting, we can accurately determine the ground truth by sampling a very large number of images (large $M$).
However, for simulated analyses with small sample sizes (small $M$) we find that the plug-in estimator of similarity shows a positive bias, in agreement to our observations in Figure~4.
In contrast, the moment-based estimator provides, on average, a better estimate of the shape similarity (albeit with higher variance across simulated analyses).
We also observe that the bias of the plug-in estimator depends on how quickly the eigenvalues of the response covariance decays. (``effective dimensionality''; see e.g. \citealt{elmoznino2022high}).
Thus, analyses of shape distance across large collections of networks risk contamination from confounding variables, such as effective dimensionality, in under-powered regimes.
This underscores the importance of removing bias, either by using sufficiently large $M$ or using an alternative approach, such as our moment-based estimator.
Overall, our observations on artificial network representations qualitatively agree with our simulated results (sec.~\ref{subsec:simulated-data-application}) and analysis of biological data (sec.~\ref{subsec:biological-data-application}).
\section{Discussion}
\label{sec:discussion}

There is a vast literature of papers that utilize or develop measures of representational similarity between neural networks~\citep[see][for review]{Klabunde2023}.
Recent works have shown interest in leveraging distances that satisfy the triangle inequality~\citep{Williams2021_shape_metrics,lange2022neural,Duong2023,Giaffar2023}, yet the statistical properties of these shape distance measures is understudied.
Here, we theoretically characterized ``plug-in'' estimates of shape distance in high-dimensional, noisy, and sample-limited regimes.
We found that these estimates \textit{(a)} tend to over-estimate representational similarity when the true similarity is small and \textit{(b)} require a large number of samples, $M$, to overcome this bias in high-dimensional regimes.
\Cref{theorem:upper-bound-on-squared-procrustes,theorem:lower-performance-bound} provide precise guarantees on the worst-case performance of plug-in estimators, which should guide the design of biological experiments and analyses of their statistical power.

An equally important contribution of our work is to provide a practical method to \textit{(a)} reduce the bias of plug-in estimators of shape distance, \textit{(b)} quantify uncertainty in shape distance estimates, and \textit{(c)} enable practicioners to explicitly trade off estimator bias and variance.
When employed on a biological dataset published by \citet{stringer2019high}, we find that shape similarity estimates are highly uncertain, revealing the challenging nature of the problem in high dimensions and with noisy data.
Importantly, this degree of uncertainty is not obvious from the procedures and plug-in estimates advertised by existing work on this subject.

Both theoretical and methodological aspects of our work may be of broader interest beyond the immediate subject of shape distance estimation.
We have seen that estimating the nuclear norm of the cross-covariance, $\Vert \mbSigma_{ij} \Vert_*$, is the key challenge in our problem.
Estimating the spectrum of cross-covariance matrices is a topic of contemporary interest~\citep{BenaychGeorges2023}, and further exploring the connections between this problem and shape distance estimation is an intriguing direction. Similarly, the method-of-moments estimator presented in \cref{subsec:method-of-moments-estimator} is broadly applicable to generalized trace estimation \citep{Adams2018}.

While others have used polynomial expansions in this context \citep{Lin2016}, a key novelty of our approach is the selection of coefficients with a tunable parameter that explicitly trades off estimator bias and variance. A more typical approach would be to choose these coefficients based on a Chebyshev polynomial expansion. While elegant, we believe our procedure for tuning these coefficients will be more relevant to scientific applications where samples are limited (such as neural data) and practitioners desire finer-scale control.

In summary, our work rigorously interrogates the statistical challenges of estimating shape distances in high-dimensional spaces.
While shape distances can be well-behaved in certain settings (e.g. in noiseless artificial networks with many sampled conditions), our theoretical results and empirical observations suggest the need for carefully designed experiments and estimation procedures.

\clearpage

\section*{Acknowledgements}

This work was supported by the Center for Computational Neuroscience at the Flatiron Institute of the
Simons Foundation. DAP was supported Simons Collaboration on the Global Brain (SCGB AWD543027). We thank Jonathan Pillow for useful feedback. The code for our project is available at \url{https://github.com/dp4846/eigmom_shape_stats/}.

\bibliography{references}

\begin{thebibliography}{34}
\providecommand{\natexlab}[1]{#1}
\providecommand{\url}[1]{\texttt{#1}}
\expandafter\ifx\csname urlstyle\endcsname\relax
  \providecommand{\doi}[1]{doi: #1}\else
  \providecommand{\doi}{doi: \begingroup \urlstyle{rm}\Url}\fi

\bibitem[Adams et~al.(2018)Adams, Pennington, Johnson, Smith, Ovadia, Patton,
  and Saunderson]{Adams2018}
Ryan~P. Adams, Jeffrey Pennington, Matthew~J. Johnson, Jamie Smith, Yaniv
  Ovadia, Brian Patton, and James Saunderson.
\newblock Estimating the spectral density of large implicit matrices, 2018.

\bibitem[Andrade et~al.(2004)Andrade, G\'{o}mez-Carracedo, Krzanowski, and
  Kubista]{andrade_procrustes_2004}
Jose~Manuel Andrade, Mar\'{i}a~P. G\'{o}mez-Carracedo, Wojtek Krzanowski, and
  Mikael Kubista.
\newblock Procrustes rotation in analytical chemistry, a tutorial.
\newblock \emph{Chemometrics and Intelligent Laboratory Systems}, 72\penalty0
  (2):\penalty0 123--132, July 2004.
\newblock ISSN 0169-7439.
\newblock \doi{10.1016/j.chemolab.2004.01.007}.
\newblock URL
  \url{https://www.sciencedirect.com/science/article/pii/S0169743904000152}.

\bibitem[Benaych-Georges et~al.(2023)Benaych-Georges, Bouchaud, and
  Potters]{BenaychGeorges2023}
Florent Benaych-Georges, Jean-Philippe Bouchaud, and Marc Potters.
\newblock {Optimal cleaning for singular values of cross-covariance matrices}.
\newblock \emph{The Annals of Applied Probability}, 33\penalty0 (2):\penalty0
  1295 -- 1326, 2023.
\newblock \doi{10.1214/22-AAP1842}.
\newblock URL \url{https://doi.org/10.1214/22-AAP1842}.

\bibitem[Boyd \& Vandenberghe(2004)Boyd and Vandenberghe]{Boyd2004-sl}
Stephen Boyd and Lieven Vandenberghe.
\newblock \emph{Convex optimization}.
\newblock Cambridge University Press, 2004.

\bibitem[Cai et~al.(2016)Cai, Schuck, Pillow, and Niv]{Cai2016}
Mingbo Cai, Nicolas~W Schuck, Jonathan~W Pillow, and Yael Niv.
\newblock A bayesian method for reducing bias in neural representational
  similarity analysis.
\newblock In D.~Lee, M.~Sugiyama, U.~Luxburg, I.~Guyon, and R.~Garnett (eds.),
  \emph{Advances in Neural Information Processing Systems}, volume~29. Curran
  Associates, Inc., 2016.
\newblock URL
  \url{https://proceedings.neurips.cc/paper/2016/file/b06f50d1f89bd8b2a0fb771c1a69c2b0-Paper.pdf}.

\bibitem[Deng et~al.(2009)Deng, Dong, Socher, Li, Li, and
  Fei-Fei]{deng2009imagenet}
Jia Deng, Wei Dong, Richard Socher, Li-Jia Li, Kai Li, and Li~Fei-Fei.
\newblock Imagenet: A large-scale hierarchical image database.
\newblock In \emph{2009 IEEE conference on computer vision and pattern
  recognition}, pp.\  248--255. Ieee, 2009.

\bibitem[Duong et~al.(2023)Duong, Zhou, Nassar, Berman, Olieslagers, and
  Williams]{Duong2023}
Lyndon~R. Duong, Jingyang Zhou, Josue Nassar, Jules Berman, Jeroen Olieslagers,
  and Alex~H. Williams.
\newblock Representational dissimilarity metric spaces for stochastic neural
  networks.
\newblock In \emph{International Conference on Learning Representations}, 2023.

\bibitem[Elmoznino \& Bonner(2022)Elmoznino and Bonner]{elmoznino2022high}
Eric Elmoznino and Michael~F Bonner.
\newblock High-performing neural network models of visual cortex benefit from
  high latent dimensionality.
\newblock \emph{bioRxiv}, pp.\  2022--07, 2022.

\bibitem[Giaffar et~al.(2023)Giaffar, Bux{\'o}, and Aoi]{Giaffar2023}
Hamza Giaffar, Camille~Rull{\'a}n Bux{\'o}, and Mikio Aoi.
\newblock The effective number of shared dimensions: A simple method for
  revealing shared structure between datasets.
\newblock \emph{bioRxiv}, 2023.
\newblock \doi{10.1101/2023.07.27.550815}.
\newblock URL
  \url{https://www.biorxiv.org/content/early/2023/07/28/2023.07.27.550815}.

\bibitem[Goodall(1991)]{Goodall1991}
Colin~R. Goodall.
\newblock Procrustes methods in the statistical analysis of shape.
\newblock \emph{Journal of the royal statistical society series
  b-methodological}, 53:\penalty0 285--321, 1991.
\newblock URL \url{https://api.semanticscholar.org/CorpusID:53315995}.

\bibitem[Gower \& Dijksterhuis(2004)Gower and
  Dijksterhuis]{gower2004procrustes}
John~C Gower and Garmt~B Dijksterhuis.
\newblock \emph{Procrustes problems}, volume~30.
\newblock OUP Oxford, 2004.

\bibitem[Harvey et~al.(2023)Harvey, Larsen, and Williams]{harvey_duality_2023}
Sarah~E. Harvey, Brett~W. Larsen, and Alex~H. Williams.
\newblock Duality of {Bures} and {Shape} {Distances} with {Implications} for
  {Comparing} {Neural} {Representations}.
\newblock November 2023.
\newblock \doi{10.48550/arXiv.2311.11436}.
\newblock URL \url{http://arxiv.org/abs/2311.11436}.
\newblock arXiv:2311.11436 [cs, stat].

\bibitem[He et~al.(2016)He, Zhang, Ren, and Sun]{he2016deep}
Kaiming He, Xiangyu Zhang, Shaoqing Ren, and Jian Sun.
\newblock Deep residual learning for image recognition.
\newblock In \emph{Proceedings of the IEEE conference on computer vision and
  pattern recognition}, pp.\  770--778, 2016.

\bibitem[Kendall et~al.(2009)Kendall, Barden, Carne, and Le]{Kendall2009-zb}
D~G Kendall, D~Barden, T~K Carne, and H~Le.
\newblock \emph{Shape and Shape Theory}.
\newblock John Wiley \& Sons, September 2009.

\bibitem[Kent \& Mardia(1997)Kent and Mardia]{Kent1997}
John~T. Kent and Kanti~V. Mardia.
\newblock Consistency of procrustes estimators.
\newblock \emph{Journal of the Royal Statistical Society. Series B
  (Methodological)}, 59\penalty0 (1):\penalty0 281--290, 1997.
\newblock ISSN 00359246.
\newblock URL \url{http://www.jstor.org/stable/2345930}.

\bibitem[Klabunde et~al.(2023)Klabunde, Schumacher, Strohmaier, and
  Lemmerich]{Klabunde2023}
Max Klabunde, Tobias Schumacher, Markus Strohmaier, and Florian Lemmerich.
\newblock Similarity of neural network models: A survey of functional and
  representational measures, 2023.

\bibitem[Kong \& Valiant(2017)Kong and Valiant]{Kong2017}
Weihao Kong and Gregory Valiant.
\newblock {Spectrum estimation from samples}.
\newblock \emph{The Annals of Statistics}, 45\penalty0 (5):\penalty0 2218 --
  2247, 2017.
\newblock \doi{10.1214/16-AOS1525}.
\newblock URL \url{https://doi.org/10.1214/16-AOS1525}.

\bibitem[Kornblith et~al.(2019)Kornblith, Norouzi, Lee, and
  Hinton]{Kornblith2019}
Simon Kornblith, Mohammad Norouzi, Honglak Lee, and Geoffrey Hinton.
\newblock Similarity of neural network representations revisited.
\newblock In Kamalika Chaudhuri and Ruslan Salakhutdinov (eds.),
  \emph{Proceedings of the 36th International Conference on Machine Learning},
  volume~97 of \emph{Proceedings of Machine Learning Research}, pp.\
  3519--3529. PMLR, 09--15 Jun 2019.
\newblock URL \url{https://proceedings.mlr.press/v97/kornblith19a.html}.

\bibitem[Kriegeskorte et~al.(2008{\natexlab{a}})Kriegeskorte, Mur, and
  Bandettini]{Kriegeskorte2008}
Nikolaus Kriegeskorte, Marieke Mur, and Peter Bandettini.
\newblock Representational similarity analysis - connecting the branches of
  systems neuroscience.
\newblock \emph{Front. Syst. Neurosci.}, 2:\penalty0 4, November
  2008{\natexlab{a}}.

\bibitem[Kriegeskorte et~al.(2008{\natexlab{b}})Kriegeskorte, Mur, Ruff, Kiani,
  Bodurka, Esteky, Tanaka, and Bandettini]{Kriegeskorte2008-su}
Nikolaus Kriegeskorte, Marieke Mur, Douglas~A Ruff, Roozbeh Kiani, Jerzy
  Bodurka, Hossein Esteky, Keiji Tanaka, and Peter~A Bandettini.
\newblock Matching categorical object representations in inferior temporal
  cortex of man and monkey.
\newblock \emph{Neuron}, 60\penalty0 (6):\penalty0 1126--1141, December
  2008{\natexlab{b}}.

\bibitem[Lange et~al.(2022)Lange, Kwok, Matelsky, Wang, Rolnick, and
  Kording]{lange2022neural}
Richard~D Lange, Devin Kwok, Jordan Matelsky, Xinyue Wang, David~S Rolnick, and
  Konrad~P Kording.
\newblock Neural networks as paths through the space of representations.
\newblock \emph{arXiv preprint arXiv:2206.10999}, 2022.

\bibitem[Lin et~al.(2016)Lin, Saad, and Yang]{Lin2016}
Lin Lin, Yousef Saad, and Chao Yang.
\newblock Approximating spectral densities of large matrices.
\newblock \emph{SIAM Review}, 58\penalty0 (1):\penalty0 34--65, 2016.
\newblock \doi{10.1137/130934283}.

\bibitem[Potters \& Bouchaud(2020)Potters and Bouchaud]{potters_bouchaud_2020}
Marc Potters and Jean-Philippe Bouchaud.
\newblock \emph{A First Course in Random Matrix Theory: for Physicists,
  Engineers and Data Scientists}.
\newblock Cambridge University Press, 2020.
\newblock \doi{10.1017/9781108768900}.

\bibitem[Raghu et~al.(2017)Raghu, Gilmer, Yosinski, and
  Sohl-Dickstein]{Raghu2017}
Maithra Raghu, Justin Gilmer, Jason Yosinski, and Jascha Sohl-Dickstein.
\newblock Svcca: Singular vector canonical correlation analysis for deep
  learning dynamics and interpretability.
\newblock In I.~Guyon, U.~Von Luxburg, S.~Bengio, H.~Wallach, R.~Fergus,
  S.~Vishwanathan, and R.~Garnett (eds.), \emph{Advances in Neural Information
  Processing Systems}, volume~30. Curran Associates, Inc., 2017.

\bibitem[Rohlf \& Slice(1990)Rohlf and Slice]{rohlf_extensions_1990}
F.~James Rohlf and Dennis Slice.
\newblock Extensions of the {Procrustes} {Method} for the {Optimal}
  {Superimposition} of {Landmarks}.
\newblock \emph{Systematic Zoology}, 39\penalty0 (1):\penalty0 40--59, 1990.
\newblock ISSN 0039-7989.
\newblock \doi{10.2307/2992207}.
\newblock URL \url{https://www.jstor.org/stable/2992207}.
\newblock Publisher: [Oxford University Press, Society of Systematic
  Biologists, Taylor \& Francis, Ltd.].

\bibitem[Saito et~al.(2015)Saito, Fonseca-Gessner, and
  Siqueira]{saito_how_2015}
Victor~S. Saito, Alaide~A. Fonseca-Gessner, and Tadeu Siqueira.
\newblock How {Should} {Ecologists} {Define} {Sampling} {Effort}? {The}
  {Potential} of {Procrustes} {Analysis} for {Studying} {Variation} in
  {Community} {Composition}.
\newblock \emph{Biotropica}, 47\penalty0 (4):\penalty0 399--402, 2015.
\newblock ISSN 0006-3606.
\newblock URL \url{https://www.jstor.org/stable/48574958}.
\newblock Publisher: [Association for Tropical Biology and Conservation,
  Wiley].

\bibitem[Schütt et~al.(2023)Schütt, Kipnis, Diedrichsen, and
  Kriegeskorte]{Schutt2023}
Heiko~H Schütt, Alexander~D Kipnis, Jörn Diedrichsen, and Nikolaus
  Kriegeskorte.
\newblock Statistical inference on representational geometries.
\newblock \emph{eLife}, 12:\penalty0 e82566, aug 2023.
\newblock ISSN 2050-084X.
\newblock \doi{10.7554/eLife.82566}.
\newblock URL \url{https://doi.org/10.7554/eLife.82566}.

\bibitem[Shi et~al.(2019)Shi, Shea-Brown, and Buice]{Shi2019}
Jianghong Shi, Eric Shea-Brown, and Michael Buice.
\newblock Comparison against task driven artificial neural networks reveals
  functional properties in mouse visual cortex.
\newblock In H.~Wallach, H.~Larochelle, A.~Beygelzimer, F.~d\textquotesingle
  Alch\'{e}-Buc, E.~Fox, and R.~Garnett (eds.), \emph{Advances in Neural
  Information Processing Systems}, volume~32. Curran Associates, Inc., 2019.
\newblock URL
  \url{https://proceedings.neurips.cc/paper_files/paper/2019/file/748d6b6ed8e13f857ceaa6cfbdca14b8-Paper.pdf}.

\bibitem[Stringer et~al.(2019)Stringer, Pachitariu, Steinmetz, Carandini, and
  Harris]{stringer2019high}
Carsen Stringer, Marius Pachitariu, Nicholas Steinmetz, Matteo Carandini, and
  Kenneth~D Harris.
\newblock High-dimensional geometry of population responses in visual cortex.
\newblock \emph{Nature}, 571\penalty0 (7765):\penalty0 361--365, 2019.

\bibitem[Tropp(2015)]{Tropp2015}
Joel~A. Tropp.
\newblock An introduction to matrix concentration inequalities.
\newblock \emph{Foundations and Trends® in Machine Learning}, 8\penalty0
  (1-2):\penalty0 1--230, 2015.
\newblock ISSN 1935-8237.
\newblock \doi{10.1561/2200000048}.
\newblock URL \url{http://dx.doi.org/10.1561/2200000048}.

\bibitem[Wainwright(2019)]{wainwright2019high}
Martin~J Wainwright.
\newblock \emph{High-dimensional statistics: A non-asymptotic viewpoint},
  volume~48.
\newblock Cambridge university press, 2019.

\bibitem[Walther et~al.(2016)Walther, Nili, Ejaz, Alink, Kriegeskorte, and
  Diedrichsen]{Walther2016}
Alexander Walther, Hamed Nili, Naveed Ejaz, Arjen Alink, Nikolaus Kriegeskorte,
  and J{\"o}rn Diedrichsen.
\newblock Reliability of dissimilarity measures for multi-voxel pattern
  analysis.
\newblock \emph{NeuroImage}, 137:\penalty0 188--200, 2016.

\bibitem[Williams \& Linderman(2021)Williams and
  Linderman]{Williams2021_single_trial}
Alex~H. Williams and Scott~W. Linderman.
\newblock Statistical neuroscience in the single trial limit.
\newblock \emph{Current Opinion in Neurobiology}, 70:\penalty0 193--205, 2021.
\newblock ISSN 0959-4388.
\newblock \doi{https://doi.org/10.1016/j.conb.2021.10.008}.
\newblock URL
  \url{https://www.sciencedirect.com/science/article/pii/S0959438821001203}.
\newblock Computational Neuroscience.

\bibitem[Williams et~al.(2021)Williams, Kunz, Kornblith, and
  Linderman]{Williams2021_shape_metrics}
Alex~H Williams, Erin Kunz, Simon Kornblith, and Scott Linderman.
\newblock Generalized shape metrics on neural representations.
\newblock In M.~Ranzato, A.~Beygelzimer, Y.~Dauphin, P.S. Liang, and J.~Wortman
  Vaughan (eds.), \emph{Advances in Neural Information Processing Systems},
  volume~34, pp.\  4738--4750. Curran Associates, Inc., 2021.
\newblock URL
  \url{https://proceedings.neurips.cc/paper/2021/file/252a3dbaeb32e7690242ad3b556e626b-Paper.pdf}.

\end{thebibliography}
\bibliographystyle{iclr2024_conference}
\clearpage

\appendix
\section{Appendix: Background on Generalized Shape Metrics}
Here we provide several relevant derivations for generalized shape metrics.
For a more thorough review, we direct the reader to \cite{Williams2021_shape_metrics} for the foundational results on generalized shape metrics and \cite{Duong2023} for the extension to stochastic neural networks.

We can intuitively think of the Procrustes distance as the Euclidean distance between two vectors remaining when the rotations and reflections have been ``removed". 
Similarly, the Riemannian shape distance can be thought of as the angle between two vectors after these rotations and reflections are removed.  
These definitions in \cref{eq:procrustes} and \cref{eq:riemannian} also make clear that Procrustes distance, like Euclidean distance, is sensitive to the overall scaling of $h_i$ or $h_j$, while the Riemannian shape distance, like the angle between vectors, is scale-invariant.

\subsection{Equivalence of \cref{eq:procrustes,eq:procrustes-alt}; \cref{eq:riemannian,eq:riemannian-alt}}
\label{app:reformulate}

The squared Procrustes can be reformulated in terms of the covariance and cross-covariance matrices as follows:
\begin{align*}
\rho^2(h_i, h_j) &= \min_{\mbQ \in \cO(N)} \expect \Vert h_i(\mbz) - \mbQ h_j(\mbz) \Vert_2^2 \\
&= \min_{\mbQ \in \cO(N)} \expect  \left [ h_i(\mbz)^\top h_i(\mbz) + h_j(\mbz)^\top h_j(\mbz) - 2 h_i(\mbz)^\top \mbQ h_j(\mbz) \right ] \\
&= \expect  \left [ h_i(\mbz)^\top h_i(\mbz) \right ] + \expect \left [ h_j(\mbz)^\top h_j(\mbz) \right ] - 2 \max_{\mbQ \in \cO(N)}\expect \left [ h_i(\mbz)^\top \mbQ h_j(\mbz) \right ] \\
&= \expect \left [ \Tr \left [ h_i(\mbz) h_i(\mbz)^\top \right ] \right] + \expect \left [ \Tr \left [ h_j(\mbz) h_j(\mbz)^\top \right ] \right ] - 2 \max_{\mbQ \in \cO(N)} \expect \left [ \Tr \left [ \mbQ h_j(\mbz)  h_i(\mbz)^\top \right ] \right ]  \\
&= \Tr \left [ \expect \left [ h_i(\mbz) h_i(\mbz)^\top \right ] \right] + \Tr \left [ \expect \left [ h_j(\mbz) h_j(\mbz)^\top \right ] \right ] - 2 \max_{\mbQ \in \cO(N)} \Tr \left [ \mbQ \expect \left [ h_j(\mbz)  h_i(\mbz)^\top \right ] \right ]  \\
&= \Tr \left [ \mbSigma_{ii} \right] + \Tr \left [ \mbSigma_{jj} \right ] - 2 \max_{\mbQ \in \cO(N)} \Tr \left [ \mbQ \mbSigma_{ij} \right ] \\
&= \Tr \left [ \mbSigma_{ii} \right ] + \Tr \left [ \mbSigma_{jj} \right ] - 2 \Vert \mbSigma_{ij} \Vert_* 
\end{align*}

Similarly for the cosine Riemannian distance:
\begin{align*}
    \cos \theta(h_i, h_j) &= \max_{\mbQ \in \cO(N)} \left ( \frac{ \expect [ h_i(\mbz)^\top \mbQ h_j(\mbz) ]}{\sqrt{\expect[h_i(\mbz)^\top h_i(\mbz)] \expect [h_j(\mbz)^\top h_j(\mbz)]}} \right ) \\
    &=  \frac{ \max_{\mbQ \in \cO(N)} \expect \left [ \Tr [\mbQ h_j(\mbz) h_i(\mbz)^\top ]\right ]}{\sqrt{\expect \left [ \Tr[h_i(\mbz) h_i(\mbz)^\top ]\right ] \expect \left [ \Tr [ h_j(\mbz) h_j(\mbz)^\top] \right ]}} \\
    &=  \frac{ \max_{\mbQ \in \cO(N)} \Tr \left [\mbQ \expect [ h_j(\mbz) h_i(\mbz)^\top ]\right ]}{\sqrt{\Tr \left [ \expect[h_i(\mbz) h_i(\mbz)^\top ]\right ] \Tr \left [ \expect [ h_j(\mbz) h_j(\mbz)^\top] \right ]}} \\
    &= \frac{ \max_{\mbQ \in \cO(N)} \Tr \left [\mbQ \mbSigma_{ij} \right ]}{\sqrt{\Tr \left [ \mbSigma_{ii} \right ] \Tr \left [ \mbSigma_{jj} \right ]}} = \frac{ \Vert \mbSigma_{ij} \Vert_*}{\sqrt{\Tr \left [ \mbSigma_{ii} \right ] \Tr \left [ \mbSigma_{jj} \right ]}}
\end{align*}

\subsection{Reformulations of the Plug-in Estimator of Procrustes distance}
\label{app:convert-plug-in}

Let $\mbz_1, \dots, \mbz_M$ denote a set of independently and identically distributed samples in the network input space. 
Then, stack the responses of network $i$ row-wise into a matrix $\mbX_i \in \reals^{M \times N}$.
Given this set up, a common definition of Procrustes distance is \citep{gower2004procrustes}:
\begin{equation}
\min_{\mbQ \in \cO(N)} \frac{1}{\sqrt{M}} \Vert \mbX_i - \mbX_j \mbQ \Vert_F
\end{equation}
Here, we have included a multiplying factor of $1/\sqrt{M}$ for reasons that will become clear shortly.
Aside from this factor, the quantity above is how \citet{Williams2021_shape_metrics} define the Procrustes distance.
Below, we show that the square of this quantity is indeed the plug-in estimator we defined in \cref{eq:procrustes-plug-in-estimator} in terms of the empirical covariance matrices:
\begin{align*}
\min_{\mbQ \in \cO(N)} \frac{1}{M} \Vert \mbX_i - \mbX_j \mbQ \Vert_F^2 &= \min_{\mbQ \in \cO(N)} \frac{1}{M} \left ( \Tr[\mbX_i^\top \mbX_i] + \Tr[\mbX_j^\top \mbX_j] - 2 \Tr[\mbX_i \mbX_j^\top \mbQ ] \right ) \\
&= \Tr \left [ \tfrac{1}{M} \mbX_i^\top \mbX_i \right ] + \Tr \left [ \tfrac{1}{M} \mbX_j^\top \mbX_j \right ] - 2 \max_{\mbQ \in \cO(N)} \Tr \left [ \tfrac{1}{M} \mbX_i \mbX_j^\top \mbQ \right ] \\
&= \Tr \left [ \hat{\mbSigma}_{ii} \right ] + \Tr \left [ \hat{\mbSigma}_{jj} \right ] - 2 \max_{\mbQ \in \cO(N)} \Tr \left [ \hat{\mbSigma}_{ij} \mbQ \right ] \\
&= \Tr \left [ \hat{\mbSigma}_{ii} \right ] + \Tr \left [ \hat{\mbSigma}_{jj} \right ] - 2 \Vert \hat{\mbSigma}_{ij} \Vert_* \\
&= \hat{\rho}^2 (h_i, h_j)
\end{align*}

\section{Appendix: Plug-in Estimator Theory}
Here we provide a number of derivations related to the behavior of the plug-in estimator for generalized shape metrics. For readers interested in further background, we suggest \cite{wainwright2019high} and \cite{Tropp2015} for an overview of classic and matrix concentration inequalities and \cite{potters_bouchaud_2020} for an overview of random matrix theory.

\subsection{Proof of \cref{lemma:tail-bound-for-covariance}}
\label{app:plug-in-covaraince}

Here we show that the plug-in estimate of the total variance $\Tr[\hat{\mbSigma}_{ii}]$ converges to the true variance $\Tr[\mbSigma_{ii}]$ exponentially fast as $M$ increases.
We begin with some algebraic manipulations:
\begin{align*}
    \left | \Tr [ \mbSigma_{ii} - \hat{\mbSigma}_{ii} ] \right | &=  \left | \Tr \left [ \mathbb{E}_{z \sim P} [h_i(\mbz_m) h_i(\mbz_m)^\top] -  {\textstyle \frac{1}{M} \!\! \sum\limits_{m=1}^M} h_i(\mbz_m) h_i(\mbz_m)^\top \right ] \right | \\
    &=  \left |  \mathbb{E}_{z \sim P} \left [\Tr  [h_i(\mbz_m) h_i(\mbz_m)^\top  ] \right ] -  {\textstyle \frac{1}{M} \!\! \sum\limits_{m=1}^M} \Tr [ h_i(\mbz_m) h_i(\mbz_m)^\top ] \right | \\
    &= \left |  \mathbb{E}_{z \sim P} \left [\Tr  [h_i(\mbz_m)^\top h_i(\mbz_m)  ] \right ] -  {\textstyle \frac{1}{M} \!\! \sum\limits_{m=1}^M} \Tr [ h_i(\mbz_m)^\top h_i(\mbz_m) ] \right | \\
    &= \left |  \mathbb{E}_{z \sim P} \left [ h_i(\mbz_m)^\top h_i(\mbz_m)  \right ] -  {\textstyle \frac{1}{M} \!\! \sum\limits_{m=1}^M}  h_i(\mbz_m)^\top h_i(\mbz_m)  \right |
\end{align*}

where we have used the property $\Tr[\Vc{x}\Vc{x}'] = \Vc{x}'\Vc{x}$ for any column vector $\Vc{x}$ in the last two lines.

The main assumption we are going to make is that the neural responses are constrained to an $\ell_2$ ball of radius $B\sqrt{N}$ or equivalently $h_i(\mbz_m)^\top h_i(\mbz_m) \leq B^2 N$ for all stimuli in the support of $P$.  Note that this is a reasonable assumption in both biological (energy constraints) and artificial neural networks (weight decay common).  

\begin{lemma}[Bounded Random Variables are Sub-Gaussian, \citet{wainwright2019high} Example 2.4] We say that a random variable $X$ with mean $\mu$ is sub-Gaussian with parameter $\sigma$ if:
\begin{equation*}
    \expect \left [e^{\lambda(X- \mu)} \right] \leq e^{\sigma^2 \lambda^2/2} \quad \text{for all $\lambda \in \mathbb{R}$}
\end{equation*}
Intuitively, this means that the tails of $X$ fall off faster than a Gaussian. Furthermore, if $X$ is mean zero and supported on the interval $[a,b]$, the $X$ is sub-Gaussian with parameter $\sigma = (b- a)/2$.
\end{lemma}

Thus our assumption implies that each term with $\frac{1}{M}h_i(\mbz_m)^\top h_i(\mbz_m)$ is sub-Gaussian with parameter $\sigma = B\sqrt{N}/M$.
We can then immediately apply the Hoeffding bound \citep[Proposition 2.5]{wainwright2019high}  to obtain:

\begin{equation}
    \bbP \left [ \left | \Tr [ \mbSigma_{ii} - \hat{\mbSigma}_{ii} ] \right | \geq t \right ] \leq 2 \exp \left [ - \frac{M t^2}{2 B^2 N} \right ]
\end{equation}

Analogously for term (B) we obtain:

\begin{equation}
    \bbP \left [ \left | \Tr [ \mbSigma_{jj} - \hat{\mbSigma}_{jj} ] \right | \geq t \right ] \leq 2 \exp \left [ - \frac{M t^2}{2 B^2 N} \right ]
\end{equation}

\subsection{Proof of \cref{lemma:upper-bound-nuc-norm}}
\label{app:upper-cross-covariance}

Our main tool is the matrix Bernstein inequality, given as theorem 6.1.1 in \citet{Tropp2015}.
We paraphrase a version of the theorem here to keep our narrative self-contained.
\begin{theorem}[Matrix Bernstein]
\label{theorem:matrix-bernstein}
Consider a finite sequence $\{\mbS_1, \dots, \mbS_M\}$ of independent, random $N \times N$ matrices. Assume that:
\begin{equation}
\label{eq:matrix-bernstein-bounded-assumpution}
\expect \big [ \mbS_m \big ] = \mb{0} \quad \text{and} \quad \Vert \mbS_m \Vert_\infty \leq L \quad \text{for each index } m
\end{equation}
where $\Vert \mbS_m \Vert_\infty = \sup \{ \Vert \mbS_m \mbv \Vert_2 \, : \, \Vert \mbv \Vert_2 \leq 1 \}$ is the matrix operator norm.

Further, define the variance of the sum $\sum_m \mbS_m$ as:
\begin{equation}
\label{eq:matrix-bernstein-variance-definition}
V = \left\Vert {\textstyle\sum_{m}} \expect \mbS_m^\top \mbS_m \right \Vert_\infty = \left \Vert {\textstyle\sum_{m}} \expect \mbS_m \mbS_m^\top \right \Vert_\infty
\end{equation}
Then:
\begin{equation}
\expect \big [ \, \big \Vert \,  {\textstyle\sum_{m}} \mbS_m \, \big \Vert_\infty \, \big ] \leq \sqrt{2 V \log (2N)} + \frac{L}{3} \log(2 N)
\end{equation}
\end{theorem}

We now turn to the proof of \cref{theorem:upper-bound-on-squared-procrustes}.
Define:
\begin{equation}
\mbS_m = \frac{1}{M} \left ( h_i(\mbz_m) h_j(\mbz_m)^\top - \mbSigma_{ij} \right )
\end{equation}
for the sequence of network inputs $\{\mbz_1 , \dots, \mbz_M \}$.
Notice that:
\begin{equation}
\expect \big [ \mbS_m \big ] = \frac{1}{M} \left ( \expect \left [ h_i(\mbz_m) h_j(\mbz_m)^\top \right ] - \mbSigma_{ij} \right ) = \frac{1}{M} \left ( \mbSigma_{ij} - \mbSigma_{ij} \right ) = \mb{0}
\end{equation}
Next, due to triangle inequality, we have:
\begin{equation}
\left \Vert \mbS_m \right \Vert_\infty = \frac{1}{M} \left \Vert h_i(\mbz_m) h_j(\mbz_m)^\top - \mbSigma_{ij}
\right \Vert_\infty \leq \frac{1}{M} \underbrace{\left \Vert h_i(\mbz_m) h_j(\mbz_m)^\top \right \Vert_\infty}_{(1)} + \frac{1}{M} \underbrace{\left \Vert \mbSigma_{ij}
\right \Vert_\infty}_{(2)}
\end{equation}
Terms (1) and (2) are each upper bounded by $B^2 N$, since for term (1):
\begin{align}
\left \Vert h_i(\mbz_m) h_j(\mbz_m)^\top 
\right \Vert_\infty &\leq \left \Vert h_i(\mbz_m) h_j(\mbz_m)^\top \mbv \right \Vert_2 && \text{(for any vector $\Vert \mbv \Vert_2 \leq 1$)} \\
&=   h_j(\mbz_m)^\top \mbv \left \Vert  h_i(\mbz_m) \right \Vert_2 \\
&\leq  \Vert h_j(\mbz_m) \Vert_2 \left \Vert \mbv \right \Vert_2 \left \Vert  h_i(\mbz_m) \right \Vert_2 && \text{(Cauchy-Schwarz inequality)} \\
&\leq B \sqrt{N} \cdot 1 \cdot B\sqrt{N} = B^2 N && \text{(From assumptions in eq.~\ref{eq:constraints})} 
\end{align}
And for term (2):
\begin{align}
\left \Vert \mbSigma_{ij} \right \Vert_\infty
&= \left \Vert \expect \, h_i(\mbz) h_j(\mbz)^\top \right \Vert_\infty \\
&\leq \left \Vert \expect \, h_i(\mbz_m) h_j(\mbz_m)^\top \mbv \right \Vert_2 && \text{(for any vector $\Vert \mbv \Vert_2 \leq 1$)} \\
&\leq \expect \left \Vert h_i(\mbz_m) h_j(\mbz_m)^\top \mbv \right \Vert_2  && \text{(Jensen's inequality)} \\
&\leq  B^2 N && \text{(Repeat the upper bound on term 1)} 
\end{align}
To summarize, we have:
\begin{equation}
\left \Vert \mbS_m \right \Vert_\infty \leq \frac{1}{M} \left \Vert h_i(\mbz_m) h_j(\mbz_m)^\top \right \Vert_\infty + \frac{1}{M} \left \Vert \mbSigma_{ij}
\right \Vert_\infty \leq \frac{2 B^2 N}{M}
\end{equation}
That is, we have shown that the assumptions of \cref{eq:matrix-bernstein-bounded-assumpution} are satisfied with $L = 2 B^2 N / M$.

Our next task is to determine an expression for the variance $V$ defined in \cref{eq:matrix-bernstein-variance-definition}.
First, we have:
\begin{align*}
\expect \, \mbS_m^\top \mbS_m &=\frac{1}{M^2}\expect [ h_j(\mbz_m) h_i(\mbz_m)^\top h_i(\mbz_m) h_j(\mbz_m)^\top + \mbSigma_{ij}^\top \mbSigma_{ij} - \mbSigma_{ij}^\top h_j(\mbz_m) h_i(\mbz_m)^\top - h_i(\mbz_m) h_j(\mbz_m)^\top \mbSigma_{ij} ] \\
&=\frac{1}{M^2}\expect [ h_j(\mbz_m) h_i(\mbz_m)^\top h_i(\mbz_m) h_j(\mbz_m)^\top ] + \mbSigma_{ij}^\top \mbSigma_{ij} - \mbSigma_{ij}^\top \expect [ h_j(\mbz_m) h_i(\mbz_m)^\top] - \expect [h_i(\mbz_m) h_j(\mbz_m)^\top] \mbSigma_{ij} \\
&=\frac{1}{M^2}\expect [ h_j(\mbz_m) h_i(\mbz_m)^\top h_i(\mbz_m) h_j(\mbz_m)^\top ] + \mbSigma_{ij}^\top \mbSigma_{ij} - \mbSigma_{ij}^\top \mbSigma_{ij} - \mbSigma_{ij}^\top \mbSigma_{ij} \\
&=\frac{1}{M^2}\expect [ h_j(\mbz_m) h_i(\mbz_m)^\top h_i(\mbz_m) h_j(\mbz_m)^\top ] - \mbSigma_{ij}^\top \mbSigma_{ij}
\end{align*}
Then, by triangle inequality:
\begin{align*}
\Vert \expect \, \mbS_m^\top \mbS_m \Vert_\infty &= \frac{1}{M^2} \Vert \expect [ h_j(\mbz_m) h_i(\mbz_m)^\top h_i(\mbz_m) h_j(\mbz_m)^\top ] - \mbSigma_{ij}^\top \mbSigma_{ij} \Vert_\infty \\
&\leq \frac{1}{M^2} \underbrace{\Vert \expect [ h_j(\mbz_m) h_i(\mbz_m)^\top h_i(\mbz_m) h_j(\mbz_m)^\top ] \Vert_\infty}_{\text{(A)}} + \frac{1}{M^2} \underbrace{\Vert \mbSigma_{ij}^\top \mbSigma_{ij} \Vert_\infty}_{\text{(B)}}
\end{align*}
Terms (A) and (B) are each upper bounded by $N^2$. 
First, taking term (A):
\begin{align*}
\left \Vert \expect \, [ h_j(\mbz_m) h_i(\mbz_m)^\top h_i(\mbz_m) h_j(\mbz_m)^\top ] \right \Vert_\infty & \leq \left \Vert \expect \, [ h_j(\mbz_m) h_i(\mbz_m)^\top h_i(\mbz_m) h_j(\mbz_m)^\top \mbv ] \right \Vert_2  && \text{(for $\Vert \mbv \Vert \leq 1$)} \\
&\leq \expect \left \Vert h_j(\mbz_m) h_i(\mbz_m)^\top h_i(\mbz_m) h_j(\mbz_m)^\top \mbv \right \Vert_2 && \text{(Jensen's)} \\
&\leq \expect \, \big [ h_j(\mbz_m)^\top \mbv \Vert h_i(\mbz_m) \Vert_2^2 \left \Vert h_j(\mbz_m) \right \Vert_2 \big ] \\
&\leq \expect \, \big [ \Vert \mbv \Vert_2  \Vert h_i(\mbz_m) \Vert_2^2 \left \Vert h_j(\mbz_m) \right \Vert_2^2 \big ] && \text{(Cauchy-Schwarz)} \\
&\leq 1 \cdot B^2 N \cdot B^2 N = B^4 N^2 && \text{(from eq.~\ref{eq:constraints})}
\end{align*}
For term (B), we first note that ${\Vert \mbSigma_{ij}^\top \mbSigma_{ij} \Vert_\infty \leq \Vert \mbSigma_{ij} \Vert_\infty^2}$ due to the fact that the operator norm is submultiplicative.
Then, term (B) is upper bounded by $B^4 N^2$ follows readily from:
\begin{align*}
\Vert \mbSigma_{ij} \Vert_\infty &= \Vert \expect \, h_i(\mbz) h_j(\mbz)^\top \Vert_\infty \\
&\leq \Vert \expect \, h_i(\mbz) h_j(\mbz)^\top \mbv \Vert_2 && \text{(for $\Vert \mbv \Vert \leq 1$)} \\
&\leq \expect \,  \Vert h_i(\mbz) h_j(\mbz)^\top \mbv \Vert_2 && \text{(Jensen's)} \\
&\leq \expect \,  \Vert h_i(\mbz) \Vert_2 \Vert h_j(\mbz) \Vert_2 \Vert \mbv \Vert_2  && \text{(Cauchy-Schwarz)} \\
&\leq B\sqrt{N} \cdot B\sqrt{N} \cdot 1 = B^2N && \text{(from eq. 1)}
\end{align*}
Taking these two bounds together, we have shown $\Vert \expect \, \mbS_m^\top \mbS_m \Vert_\infty \leq 2 B^4 N^2 / M^2$.
We are now ready to upper bound the variance term, $V$, appearing in \cref{theorem:matrix-bernstein}.
Specifically, by the triangle inequality and the bounds above, we have:
\begin{equation}
V = \Vert {\textstyle\sum_m} \expect \, \mbS_m^\top \mbS_m^\top \Vert_\infty \leq \sum_{m=1}^M \Vert \expect \, \mbS_m^\top \mbS_m^\top \Vert_\infty \leq \frac{2 B^4 N^2}{M}
\end{equation}

With this, we are equipped to apply the matrix Bernstein inequality to obtain an upper bound on the estimation error of the plug-in estimator.
Specifically, we have:
\begin{align*}
\left | \Vert \hat{\mbSigma}_{ij} \Vert_* - \Vert \mbSigma_{ij} \Vert_* \right | &\leq  \Vert \hat{\mbSigma}_{ij} - \mbSigma_{ij} \Vert_*  && \text{(reverse triangle inequality)}\\
&= \Vert \sum_m \mbS_m \Vert_* \\
&\leq N \Vert \sum_m \mbS_m \Vert_\infty
\\
&\leq N \sqrt{2 V \log (2N)} + \frac{NL}{3} \log(2 N)
&& \text{(\cref{theorem:matrix-bernstein})} \\
&\leq 2 B^2 N^2 M^{-1/2}\sqrt{\log(2N)} + \frac{2 B^2 N^2}{3M} \log(2 N)
\end{align*}
Where we have substituted the derived quantities $L = 2 B^2 N / M$ and $V \leq 2 B^4 N^2 / M$ in the final line.

\subsection{Proof of \cref{theorem:upper-bound-on-squared-procrustes}}
\label{app:tail-cross-covariance}

\Cref{lemma:upper-bound-nuc-norm} provides an upper bound on the expected value on $\left | \Vert \mbSigma_{ij} \Vert_* - \Vert \hat{\mbSigma}_{ij} \Vert_*  \right |$, which is the error of our plug-in estimate of cross-covariance nuclear norm.
This bound holds for any true cross-covariance matrix $\mbSigma_{ij}$, provided that the constraints in \cref{eq:constraints} are satisfied.
However, this tells us nothing about how the estimation error deviates around its expectation.

Here, we use the bounded differences inequality \citep[Corollary 2.21]{wainwright2019high}, also called McDiarmid's inequality, to show that deviations around this expectation decrease exponentially fast.
Thus, the upper bound on the expected error (\cref{theorem:upper-bound-on-squared-procrustes}) provides accurate intuition.

\begin{lemma}[Bounded Differences Inequality, \citet{wainwright2019high} Corollary 2.21] Consider a function $f: \mathbb{R}^n \rightarrow \mathbb{R}$.  The function is said to have the bounded difference property for the $k$th coordinate if there exists an $L_k$ for which the following holds:
\begin{equation*}
    \max_{X_{1:n} \in \mathbb{R}^n, X'_k \in \mathbb{R}} \big | f(X_{1:n}) - f(X_{1:k-1}, X'_k, X_{k+1:n} ) \big | \leq L_k
\end{equation*}
Suppose $f$ satisfies this property with $L_1, \ldots, L_n$ for each coordinate respectively.  Then the following inequality holds:
\begin{equation}
    \mathbb{P} \left [ \bigg | f(X_{1:n}) - \expect [f(X_{1:n})] \bigg | \geq t\right ] \leq \exp \left [-\frac{2t^2}{\sum_{i = 1}^n L_i^2} \right ]
\end{equation}
\end{lemma}

We start by applying the reverse triangle inequality:

\begin{align*}
    \left | \Vert \mbSigma_{ij} \Vert_* - \Vert \hat{\mbSigma}_{ij} \Vert_*  \right | & \leq \Vert \mbSigma_{ij} -  \hat{\mbSigma}_{ij} \Vert_* = \left \Vert \mbSigma_{ij} -  {\textstyle \frac{1}{M} \!\! \sum\limits_{m=1}^M}  h_i(\mbz_m) h_j(\mbz_m)^\top \right \Vert_* 
\end{align*}

We can bound how much this changes if we change one coordinate of the function, i.e. if $h_i(\mbz_1)^\top h_j(\mbz_1)$ is replaced by $h_i(\Vc{\tilde{z}}_1)^\top h_j(\Vc{\tilde{z}}_1)$.  The difference is then bounded by:

\begin{gather*}
    \left \Vert \mbSigma_{ij} -  {\textstyle \frac{1}{M} \!\! \sum\limits_{m=1}^M}  h_i(\mbz_m) h_j(\mbz_m)^\top \right \Vert_* - \left \Vert \mbSigma_{ij} -  \left ( {\textstyle \frac{1}{M} \!\! \sum\limits_{m=1}^M}  h_i(\mbz_m) h_j(\mbz_m)^\top - \frac{1}{M} h_i(\mbz_1) h_j(\mbz_1)^\top + \frac{1}{M} h_i(\Vc{\tilde{z}}_1) h_j(\Vc{\tilde{z}}_1)^\top \right )  \right \Vert_* \\
    \leq  \left \Vert \frac{1}{M} \left ( h_i(\mbz_1) h_j(\mbz_1)^\top -  h_i(\Vc{\tilde{z}}_1) h_j(\Vc{\tilde{z}}_1)^\top \right ) \right \Vert_* = \frac{1}{M} \left \Vert h_i(\mbz_1) h_j(\mbz_1)^\top -  h_i(\Vc{\tilde{z}}_1) h_j(\Vc{\tilde{z}}_1)^\top \right \Vert_* \quad \quad \:\\
    \leq \frac{1}{M} \left ( \left \Vert h_i(\mbz_1) h_j(\mbz_1)^\top \right \Vert_* + \left \Vert  h_i(\Vc{\tilde{z}}_1) h_j(\Vc{\tilde{z}}_1)^\top \right \Vert_* \right ) = \frac{1}{M} \left ( \left | h_i(\mbz_1)^\top h_j(\mbz_1) \right | + \left |  h_i(\Vc{\tilde{z}}_1)^\top h_j(\Vc{\tilde{z}}_1) \right | \right )
\end{gather*}

Finally, we can apply Cauchy-Schwartz and our assumption about the neural activations being bounded to obtain:

\begin{align*}
    \frac{1}{M} \left ( \left | h_i(\mbz_1)^\top h_j(\mbz_1) \right | + \left |  h_i(\Vc{\tilde{z}}_1)^\top h_j(\Vc{\tilde{z}}_1) \right | \right ) &\leq \frac{1}{M} \left ( \| h_i(\mbz_1)\|_2 \|h_j(\mbz_1) \|_2 + \|  h_i(\Vc{\tilde{z}}_1) \|_2  \| h_j(\Vc{\tilde{z}}_1) \|_2 \right ) \\
    &\leq \frac{2 B^2 N}{M}
\end{align*}

Thus we have $\sum_{i=1}^M L_i^2 = \sum_{i=1}^M 4B^4 N^2/M^2 = 4B^4 N^2/M$, and we can apply the bounded differences inequality to obtain for all $t \geq 0$:

\begin{equation}
    \bbP \bigg [ \, \bigg | \left | \Vert \mbSigma_{ij} \Vert_* - \Vert \hat{\mbSigma}_{ij} \Vert_* \right |  - \expect \left | \Vert \mbSigma_{ij} \Vert_* - \Vert \hat{\mbSigma}_{ij} \Vert_* \right | \bigg | \geq t \bigg ] \leq 2 \exp \left [ - \frac{Mt^2}{2B^4N^2} \right ]
\end{equation}

For the deviation from the expectation to be in the range $[-t,t]$ with probability $1 - \delta$ we require:
\begin{equation*}
    2 \exp \left [ - \frac{Mt^2}{2B^4 N^2} \right ] \leq \delta
\end{equation*}

Solving for $t$ gives $t \geq B^2 N  M^{-1/2} \sqrt{2 \log \left (2/\delta \right )}$, and thus with probability $1- \delta$ the following holds:
\begin{equation*}
    \bigg | \left | \Vert \mbSigma_{ij} \Vert_* - \Vert \hat{\mbSigma}_{ij} \Vert_* \right |  - \expect \left | \Vert \mbSigma_{ij} \Vert_* - \Vert \hat{\mbSigma}_{ij} \Vert_* \right | \bigg | \leq  B^2 N M^{-1/2} \sqrt{2 \log (2/\delta)}
\end{equation*}

To proceed we break this we use a basic identity of the absolute value: if ${|a - b| < c}$ then ${a - b < c}$ and also ${b - a < c}$.
Thus, with probability at least $1 - \delta$, we have:
\begin{align*}
     \Vert \mbSigma_{ij} \Vert_* - \Vert \hat{\mbSigma}_{ij} \Vert_* &\leq \expect \left | \Vert \mbSigma_{ij} \Vert_* - \Vert \hat{\mbSigma}_{ij} \Vert_* \right | + \frac{B^2 N}{M^{1/2}} \sqrt{2 \log (2/\delta)} \\
     &\leq \frac{2 B^2 N^2}{M^{1/2}} \sqrt{\log(2N)} + \frac{2 B^2 N^2 }{3M} \log(2 N) + \frac{B^2 N}{M^{1/2}} \sqrt{2 \log (2/\delta)}
\end{align*}
And we also have with probability at least $1-\delta$, we have:
\begin{align*}
     \Vert \hat{\mbSigma}_{ij} \Vert_* - \Vert \mbSigma_{ij} \Vert_*  &\leq \expect \left | \Vert \mbSigma_{ij} \Vert_* - \Vert \hat{\mbSigma}_{ij} \Vert_* \right | + \frac{B^2 N}{M^{1/2}} \sqrt{2 \log (2/\delta)} \\
     &\leq \frac{2 B^2 N^2}{M^{1/2}} \sqrt{\log(2N)} + \frac{2 B^2 N^2 }{3M} \log(2 N) + \frac{B^2 N}{M^{1/2}} \sqrt{2 \log (2/\delta)}
\end{align*}
In the final inequalities above, we have simply plugged in our expectation bound from \cref{lemma:upper-bound-nuc-norm}.
The relations above imply that the following holds with probability $1-\delta$:
\begin{equation}
    \left | \Vert \mbSigma_{ij} \Vert_* - \Vert \hat{\mbSigma}_{ij} \Vert_* \right | \leq \frac{2N^2 \log(2 B^2 N)}{3M} + \frac{2 B^2 N^2 \sqrt{\log(2N)}}{M^{1/2}} + \frac{B^2 N}{M^{1/2}} \sqrt{2 \log \left ( \frac{2}{\delta} \right )}
\end{equation}

To complete the proof we need to combine the above tail bound with \cref{lemma:tail-bound-for-covariance}.
By the triangle inequality we have

\begin{align*}
| \hat{\rho}^2 - \rho^2 | &= \left | \Tr[\hat{\mbSigma}_{ii}] + \Tr[\hat{\mbSigma}_{jj}] - 2 \Vert \hat{\mbSigma}_{ij} \Vert_* - \Tr[\hat{\mbSigma}_{ii}] - \Tr[\hat{\mbSigma}_{jj}] + 2 \Vert \hat{\mbSigma}_{ij} \Vert_* \right | \\
&= \left | \Tr[\hat{\mbSigma}_{ii}] - \Tr[\mbSigma_{ii}] + \Tr[\hat{\mbSigma}_{jj}] - \Tr[\mbSigma_{jj}] + 2 \Vert \mbSigma_{ij} \Vert_* - 2 \Vert \hat{\mbSigma}_{ij} \Vert_*  \right | \\
&\leq \left | \Tr[\hat{\mbSigma}_{ii}] - \Tr[\mbSigma_{ii}] \right | + \left | \Tr[\hat{\mbSigma}_{jj}] - \Tr[\mbSigma_{jj}] \right | + 2 \left | \Vert \mbSigma_{ij} \Vert_* - \Vert \hat{\mbSigma}_{ij} \Vert_*  \right |
\end{align*}

Setting $\delta' = \delta/3$ in our results for these three terms yields that the following three inequalities independently hold with probability $\delta/3$:

\begin{align*}
    \left | \Tr[\mbSigma_{ii}] - \Tr [ \hat{\mbSigma}_{ii} ] \right | &\geq B N^{1/2}M^{-1/2} \sqrt{2\log(6/\delta)} \\
    \left | \Tr[\mbSigma_{jj}] - \Tr [ \hat{\mbSigma}_{jj} ] \right | &\geq B N^{1/2}M^{-1/2} \sqrt{2\log(6/\delta)} \\
    \left | \Vert \mbSigma_{ij} \Vert_* - \Vert \hat{\mbSigma}_{ij} \Vert_* \right | &\geq \frac{2N^2 \log(2 B^2 N)}{3M} + \frac{2 B^2 N^2 \sqrt{\log(2N)}}{M^{1/2}} + \frac{B^2 N}{M^{1/2}} \sqrt{2 \log \left ( \frac{6}{\delta} \right )}
\end{align*}

By applying the union bound, we obtain that all three inequalities hold simultaneously with probability $\leq \delta/3 + \delta/3 + \delta/3 = \delta$.  The three reverse inequalities then hold simultaneously with probability greater than or equal to $1 - \delta$.  Thus with probability at least $1 - \delta$, the following holds:

\begin{equation*}
| \hat{\rho}^2 - \rho^2 | \leq \frac{2 B^2 N^2 \log(2N)}{3M} + \frac{2 B^2 N^2 \sqrt{\log(2N)}}{M^{1/2}} + \left ( \frac{N B^2}{M^{1/2}} + \frac{2N^{1/2}B}{M^{1/2}} \right ) \! \! \sqrt{2 \log \left ( \frac{6}{\delta} \right )}
\end{equation*}

as claimed in \cref{theorem:upper-bound-on-squared-procrustes}.

\subsection{Proof of \cref{theorem:lower-performance-bound} (Lower Bound on Plug-In Estimator Error)}
\label{app:lower-cross-covaraince}

We derive a lower bound by constructing an explicit example where the plug-in estimator performs badly.
Specifically, we consider a scenario where two networks have entirely decorrelated, high-variance representations.
To do this, we use \textit{Rademacher random variables}---a random variable $R$ is called a Rademacher variable if it behaves as follows:
\begin{equation}
R = \begin{cases}
+1 & \text{with probability }1/2 \\
-1 & \text{with probability }1/2
\end{cases}
\end{equation}
Now, suppose we sample $M$ network inputs, $\mbz_1, \dots, \mbz_M \sim P$, independently.
Further, let $B > 0$ be the constant appearing in \cref{eq:constraints}.
For $m \in \{1, \dots, M\}$ define
\begin{equation}
X_m = \frac{1}{B} h_i(\mbz_m) \quad \text{and} \quad Y_m = \frac{1}{B}  h_j(\mbz_m)
\end{equation}
Note that $X_m$ and $Y_m$ are $N$-dimensional random vectors.
Due to \cref{eq:constraints}, we have $\Vert h_i(\mbz) \Vert_2 \leq B \sqrt{N}$ and $\Vert h_j(\mbz) \Vert_2 \leq B \sqrt{N}$ almost surely.
Thus, $\Vert X_m \Vert \leq \sqrt{N}$ and $\Vert Y_m \Vert \leq \sqrt{N}$ almost surely.

Define $X = (1/B) h_i(\mbz)$ and $Y = (1/B) h_j(\mbz)$ for randomly sampled $\mbz \sim P$.
The case we will consider is that $X$ and $Y$ are each composed of $N$ independent Rademacher variables.
One trivial way to construct this is to suppose each $\mbz \sim P$ is a random vector with $2N$ elements, all of which are independent Rademacher variables scaled by a factor $B > 0$.
Then, let $h_i : \reals^{2N} \mapsto \reals^N$ be the function which extracts the first $N$ elements of $\mbz$ and let $h_j : \reals^{2N} \mapsto \reals^N$ be the function which extracts the final $N$ elements.

Thus, we have constructed a setting where $X_1, \dots, X_M, Y_1, \dots, Y_M$ are all composed of independent Rademacher variables.
In this setting, the squared Procrustes distance is given by:
\begin{align}
\rho^2 &= \Tr[\mbSigma_{ii}] + \Tr[\mbSigma_{jj}] - 2 \Vert \mbSigma_{ij} \Vert_* \\
&= \Tr[\expect [h_i(\mbz) h_i(\mbz)^\top]] + \Tr[\expect [h_j(\mbz) h_j(\mbz)^\top]] - 2 \Vert \expect [h_i(\mbz) h_j(\mbz)^\top] \Vert_* \\
&= B^2 \cdot \left ( \Tr[\expect [X X^\top]] + \Tr[\expect [Y Y^\top]] - 2 \Vert \expect [X Y^\top] \Vert_* \right ) \\
&= B^2 \cdot \left ( \expect [X^\top X] + \expect [Y^\top Y] - 2 \Vert \expect [X]  \expect[Y^\top] \Vert_* \right ) \\
&= B^2 \cdot \left ( N + N - 0 \right ) \\
&= 2 B^2 N
\end{align}
where we have used the fact that $X$ and $Y$ are independent, mean zero, random vectors to conclude that the cross covariance is an $N \times N$ matrix filled with zeros.
Furthermore, note that $X_m^\top X_m = N$ and $Y_m^\top Y_m = N$ almost surely for all $m \in 1, \dots, M$ since they are comprised of $N$ Rademacher variables.
Thus, the plug-in estimate of the squared Procrustes distance takes the form:
\begin{align}
\hat{\rho}^2 
&= B^2 \cdot \left ( \Tr[\tfrac{1}{M}\textstyle\sum_m X_m X_m^\top] + \Tr[\frac{1}{M}\textstyle\sum_m Y_m Y_m^\top] - 2 \Vert \frac{1}{M}\textstyle\sum_m X_m Y_m^\top \Vert_* \right ) \\
&= B^2 \cdot \left ( \tfrac{1}{M} \textstyle\sum_m X_m^\top X_m + \tfrac{1}{M} \textstyle\sum_m Y_m^\top Y_m - 2 \Vert \tfrac{1}{M} \textstyle\sum_m X_m Y_m^\top \Vert_* \right ) \\
&= B^2 \cdot \left ( N + N - 2 \Vert \tfrac{1}{M} \textstyle\sum_m X_m Y_m^\top \Vert_* \right ) \\
&= 2 B^2 N - 2 B^2 \Vert \tfrac{1}{M} \textstyle\sum_m X_m Y_m^\top \Vert_*
\end{align}
Putting these two results together, we conclude that the absolute error of the plug-in estimator is:
\begin{equation}
\label{eq:rademacher-plugin-error}
| \rho^2 - \hat{\rho}^2 | = 2 B^2 \Vert \tfrac{1}{M} \textstyle\sum_m X_m Y_m^\top \Vert_*
\end{equation}

Now, the product of two indepedent Rademacher variables is also a standard Rademacher variable.
Thus, each element inside the matrix $(1/M) \sum_m X_m Y_m^\top$, is the empirical average of $M$ independent Rademacher variables.
These matrix elements are asymptotically independent in the limit that ${M \rightarrow \infty}$.
Further, the central limit theorem applies in this limit, and thus the distribution of each matrix element approaches a Gaussian distribution $\cN(0, 1/M)$.

Such random matrices are well-studied under the name of Ginibre ensembles.
In the limit that $N \rightarrow \infty$ and the variance of each matrix element is taken to be $\sigma^2 / N$, the density of the singular values takes the following form \citep[see e.g.][sec. 3.1.3]{potters_bouchaud_2020}:
\begin{equation}
    \label{eq:quarter-circle-law}
    \rho(s) = \frac{\sqrt{4 \sigma^2 - s^2}}{\pi \sigma^2} \quad s \in (0, 2 \sigma)
\end{equation}
This is called the quarter circle law since if we look at the density of $s$ it forms a quarter circle.
The nuclear norm of the matrix is $N$ times the expected value of $s$ with with respect to the density $\rho(s)$.
Integrating this density, we obtain:
\begin{align}
    \lim_{\substack{N \rightarrow \infty \\ M \gg N}} \big \Vert \tfrac{1}{M} \textstyle\sum_m X_m Y_m^\top \big \Vert_* &= \frac{N}{\pi \sigma^2} \int_{0}^{2 \sigma} s \sqrt{4 \sigma^2 - s^2} \: ds \\
    &= \frac{N}{4 \pi \sigma^2}  \left [- \frac{1}{3} (4 \sigma^2 - s^2)^{3/2} \right ]_0^{2 \sigma} \\
    &= \frac{N}{\pi \sigma^2}  \left [ \frac{1}{3} (4 \sigma^2)^{3/2} \right ] = \frac{N}{ \pi \sigma^2} \left [ \frac{8}{3} \sigma^3 \right ]\\
    &= \frac{8 \sigma}{3 \pi} N = \frac{8}{3 \pi} N^{3/2}M^{-1/2}
    \label{eq:ginibre-integrated-result}
\end{align}
Where in the last line we have substituted $\sigma = \sqrt{N / M}$, which comes from equating $\sigma^2 / N$ (the variance in of each matrix element in eq.~\ref{eq:quarter-circle-law}) with $1 / M$ (the variance given by the average of $M$ Rademacher variables under the central limit theorem).
Note that the analysis above holds asymptotically as $M, N \rightarrow \infty$ and we keep $M \gg N$ so that the central limit theorem continues to hold.

Plugging \cref{eq:ginibre-integrated-result} into \cref{eq:rademacher-plugin-error} and dividing both sides by $N$ we arrive at the expression appearing in \cref{theorem:lower-performance-bound}.

\subsection{Numerical verification of lower bound}
\label{appendix:numerical-verification-of-lower-bound}

Figure~\ref{fig:appendix-lower-bound-numerical-verification} below shows in simulation that the plug-in estimator closely agrees with the lower bound established by \cref{theorem:lower-performance-bound}.

\begin{figure}[htb]
\centering
\includegraphics[width=0.5
\linewidth]{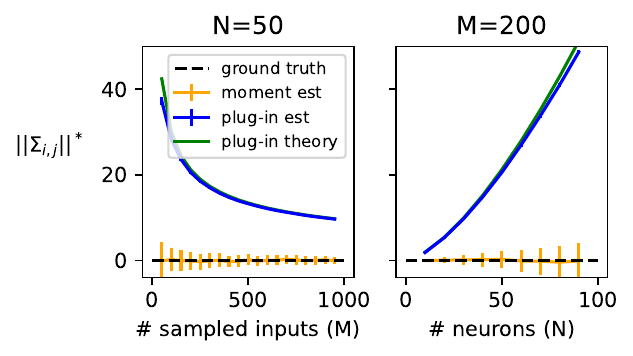}
\vspace{-0.25cm}
\caption{Simulation of lower bound on plug-in error. Simulated responses from $N$ neurons were sampled from $\cN(0, \mbI)$ independently for pairs of networks over $M$ stimuli. We plot the observed nuclear norm of the empirical cross-covariance matrix in blue, and show close agreement with equation~\ref{eq:ginibre-integrated-result} in green. The moment-based estimator is shown in yellow for comparison; note that the variance of the moment-based estimator decreases with $M$ and increases with $N$. \textit{Left}, simulations from $N=50$ neurons as $M$ was varied. \textit{Right}, simulations from $M=200$ stimuli as $N$ was varied.}
\label{fig:appendix-lower-bound-numerical-verification}
\end{figure}
\section{Appendix: Method-of-Moments Estimator}

\label{app:wp-ubiased}

\subsection{Derivation of method-of-moment estimator}

We now turn to constructing our method-of-moments estimator of $\|\Mx{\Sigma}_{ij}\|_* = \sum_{n = 1}^N s_n(\Mx{\Sigma}_{ij})$, which is required for our novel estimator of the Riemannian shape distance. We can form an unbiased estimator of the matrix $\Mx{\Sigma}_{ij}$ by observing a single random stimuli in the two networks:

\begin{equation*}
\Mx{\hat{\Sigma}}_{ijm} :=h_i(\mbz_m) h_j(\mbz_m)^{\trans} \in \bbR^{N \times N}, \quad \bbE[\Mx{\hat{\Sigma}}_{ijm}] = \Mx{\Sigma}_{ij}
\end{equation*}

Note that here the randomness comes from the selection of the stimuli, i.e. $\mbz_m \sim P$; the output of the network is deterministic. Assuming $m, m'$ are distinct stimuli drawn independently from the distribution $P$, we then have:

\begin{equation*}
    \bbE \left [ \Mx{\hat{\Sigma}}_{ijm} \Mx{\hat{\Sigma}}_{ijm'}  \right ] = \Mx{\Sigma}_{ij} \Mx{\Sigma}'_{ij}
\end{equation*}

This means we can estimate $\Mx{\Sigma}_{ij} \Mx{\Sigma}'_{ij}$ by observing a pair of stimuli in both networks.

\begin{align*}
    \Tr \left [ f( \Mx{\Sigma}_{ij}\Mx{\Sigma}'_{ij}) \right ] &= \sum_{n = 1}^N f \left ( s^2_n(\Mx{\Sigma}_{ij}) \right ) = \sum_{n = 1}^N \sum_{p=0}^{\infty} \gamma_p s^{2p}_n(\Mx{\Sigma}_{ij}) & \text{Taylor expansion of $f(\cdot)$} \\
    &=  \sum_{p=0}^{\infty} \gamma_p \sum_{n = 1}^N s^{2p}_n(\Mx{\Sigma}_{ij}) = \sum_{p=0}^{\infty} \gamma_p \Tr \left [ \left ( \Mx{\Sigma}_{ij}\Mx{\Sigma}'_{ij} \right )^p \right ] & \Tr \left [ \left ( \Mx{\Sigma}_{ij}\Mx{\Sigma}'_{ij} \right )^p \right ] = \sum_{n = 1}^N s^{2p}_n(\Mx{\Sigma}_{ij}) \\
    &= \sum_{p=0}^{\infty} \gamma_p  \bbE \left [ \Tr \left [ \prod_{\sigma = 1}^p \Mx{\hat{\Sigma}}_{ij(2\sigma-1)}\Mx{\hat{\Sigma}}'_{ij(2\sigma)} \right ] \right ] & \text{Substitute unbiased estimator for $\left ( \Mx{\Sigma}_{ij}\Mx{\Sigma}'_{ij} \right )^p$} \\
    &\approx \sum_{p=0}^{P} \gamma_p  \bbE \left [ \Tr \left [ \prod_{\sigma = 1}^p \Mx{\hat{\Sigma}}_{ij(2\sigma-1)}\Mx{\hat{\Sigma}}'_{ij(2\sigma)} \right ] \right ] & \text{Approximate with truncated power series}
\end{align*}

Our estimator for the nuclear norm of $\Mx{\Sigma}_{ij}$ is thus:

\begin{equation}
    \widehat{\|\Mx{\Sigma}_{ij}\|}_* = \sum_{p=0}^{P} \gamma_p   \Tr \left [ \prod_{\sigma = 1}^p \Mx{\hat{\Sigma}}_{ij(2\sigma-1)}\Mx{\hat{\Sigma}}'_{ij(2\sigma)} \right ]
\end{equation}

Note that for each element of the product we are considering the estimator based on stimuli $(2\sigma - 1)$ and $(2 \sigma)$; in total this estimator will use $2P$ unique stimuli.

\subsection{Deriving the Quadratic Program}
\label{app:quadratic-program}

The optimization problem in \cref{eq:gamma-quad-prog} takes the form:
\begin{equation}
\label{eq:gamma-opt-reformulation-1}
\underset{\mbgamma}{\textrm{minimize}} \quad \mbgamma^\top \mbA \mbgamma + N^2 \left ( \max_x f^2(\mbgamma, x) \right )
\end{equation}
where $f(\mbgamma, x) = x^{1/2} - \sum_p \gamma_p x^p$,
\begin{equation}
\mbgamma = \begin{bmatrix}
\gamma_1 \\ \vdots \\ \gamma_P
\end{bmatrix}
\in \reals^P , \quad
\mbA = \begin{bmatrix}
\Cov(\hat{W}_1, \hat{W}_1) & \dots & \Cov(\hat{W}_1, \hat{W}_P) \\
\vdots &  & \vdots \\
\Cov(\hat{W}_P, \hat{W}_1) & \dots & \Cov(\hat{W}_P, \hat{W}_P)
\end{bmatrix}
\in \reals^{P \times P},
\end{equation}
Notice that $f$ is linear in $\mbgamma$, and that $\mbA$ is symmetric, positive-definite.

We will reformulate \cref{eq:gamma-opt-reformulation-1} in several steps, and ultimately obtain a quadratic program that can be efficiently solved.
First, we introduce a new optimization variable $u \in \reals$ whose square is an upper bound on $f^2(\mbgamma, x)$ for all $x \in [0, 1]$.
Thus, the optimal $\mbgamma$ for the problem:
\begin{equation}
\label{eq:gamma-opt-reformulation-2}
\begin{aligned}
& \underset{\mbgamma, u}{\text{minimize}}
& & \mbgamma^\top \mbA \mbgamma + N^2 u^2 \\
& \text{subject to}
& & u^2 \geq f^2(\mbgamma, x) \quad \text{for all}~x \in [0, 1]
\end{aligned}
\end{equation}
coincides to the optimal $\mbgamma$ solving \cref{eq:gamma-opt-reformulation-1}.
This is essentially an \textit{epigraph reformulation} of the original problem \citep[see][equation 4.11]{Boyd2004-sl}.
Notice that the objective function is quadratic in this reformulation.

Next, we lay down a fine grid of linearly spaced test points $x_1, \dots, x_T \in [0, 1]$.
We can then obtain a good approximation to the solution in \cref{eq:gamma-opt-reformulation-2} by solving:
\begin{equation}
\label{eq:gamma-opt-reformulation-3}
\begin{aligned}
& \underset{\mbgamma, u}{\text{minimize}}
& & \mbgamma^\top \mbA \mbgamma + N^2 u^2 \\
& \text{subject to}
& & u^2 \geq f^2(\mbgamma, x_t) \quad \text{for all}~t \in 1, \dots, T
\end{aligned}
\end{equation}
Of course, increasing $T$ (the number of test points) improves the approximation arbitrarily well.

Finally, the constraints of the problem can be put into a form that is jointly linear in $\mbgamma$ and $u$.
First, constraining $u^2 \geq f^2(\mbgamma, x_t)$ is equivalent to simultaneously constraining ${u \geq f(\mbgamma, x_t)}$ and ${u \geq -f(\mbgamma, x_t)}$.
Then, plugging in the definition of $f(\mbgamma, x_t)$, and rearranging we have:
\begin{equation}
\label{eq:gamma-opt-reformulation-4}
\begin{aligned}
& \underset{\mbgamma, u}{\text{minimize}}
& & \mbgamma^\top \mbA \mbgamma + N^2 u^2 \\
& \text{subject to}
& & u + \sum_p \gamma_p x_t^p \geq x^{1/2}_t \quad \text{for all}~t \in 1, \dots, T \\
& & & u - \sum_p \gamma_p x_t^p \geq -x^{1/2}_t \quad \text{for all}~t \in 1, \dots, T
\end{aligned}
\end{equation}
This objective is quadratic and the constraints are linear with respect to the optimized quantities.
Thus, a solution (approximated to high accuracy) can be achieved efficiently using off-the-shelf quadratic programming solvers. To enforce the user defined bound on the bias a final two constraints are be appended to \cref{eq:gamma-opt-reformulation-4}: $-Nu \geq -c$ and $Nu \geq -c$, where $c$ is the  upper bound on the absolute bias. 

\subsection{Confidence intervals}
\label{app:confidence-intervals}
To form approximate $\alpha$ level confidence intervals around $\widehat{\|\Mx{\Sigma}_{ij}\|}_*$ we use the maximal bias (eq. \ref{eq:gamma-quad-prog}, term 1) and variance (eq. \ref{eq:gamma-quad-prog}, term 2) from the quadratic program's solution: 
$$ \left[ \widehat{\|\Mx{\Sigma}_{ij}\|}_* - z^*\sqrt{\mbgamma^\top \mbA \mbgamma} - N u , \ \  \  \  \widehat{\|\Mx{\Sigma}_{ij}\|}_* + z^*\sqrt{\mbgamma^\top \mbA \mbgamma} + N u  \right],$$

where $z^*$ is the critical value of the standard normal. For confidence intervals of the similarity score we scale this interval by the denominator of the similarity score.

\section{Appendix: Experiment Details}

\subsection{Simulated Experiments}
\label{app:simulations}

To draw data for our simulations, we set the eigenvalues of the $\Mx{\Sigma}_{ii}$ and the singular values of $\Mx{\Sigma}_{ij}$ to a ground truth nuclear norm and similarity score. To demonstrate the estimators accuracy across the space of orthogonal transformations we apply a random orthogonal rotation matrix to each population's covariance in each new parameter setting.

\subsection{Experimental data from \citet{stringer2019high}}

Neural activity in mouse primary visual cortex was recorded using a two-photon microscope while mice were free to run on an air-floating ball. Recordings were collected across multiple depth planes at a frequency of 2.5 or 3 Hz, with planes 30-35 $\mu m$ apart. The field of view of the microscope was selected such that ~10,000 neurons could be observed within a retinotopic location on the stimulus display. 

All stimuli were presented for 0.5s with a random inter-stimulus interval between 0.3 and 1.1s consisting of a grey-screen. The images used in the experiment were taken from the ImageNet database, which includes categories such as birds, cats, and insects. The researchers manually selected images that had a mix of low and high spatial frequencies and that did not consist of more than 50 \% uniform background. All images were uniformly contrast-normalized by subtracting the local mean brightness and dividing by the local mean contrast. Each stimulus consisted of a different normalized image from the ImageNet database, with 2,800 different images used in total. The same image was displayed on all three screens, but each screen showed the image at a different rotation. Each of the 2,800 natural image stimuli were displayed twice in a recording in two blocks of the same randomized order.

Calcium movie data was processed using the Suite2p toolbox to estimate spike rates of neurons. Underlying neural activity was estimated using non-negative spike deconvolution (Frierich et. al., 2017). These deconvolved traces were normalized to the mean and standard deviation of their activity during a 30-minute period of grey-screen spontaneous activity. For further detail please see the original study \cite{stringer2019high}. All analyses done in this paper were performed on the pre-processed data available on figshare (\url{https://figshare.com/articles/Recordings_of_ten_thousand_neurons_in_visual_cortex_in_response_to_2_800_natural_images/6845348}).
\section{Appendix: Applications to Deep Learning}
\label{appendix:artificial-neural-net}

Here we apply the plug-in and moment based estimator to neural network responses to demonstrate impacts of the differences between these estimators and relevance to neural networks. We make the point that the bias of the plug-in estimator, but not the moment estimator, is substantial for small samples. Furthermore, plug-in bias depends on the effective dimensionality of the two populations. Thus, naively using the plug-in can lead to erroneous scientific conclusions because the estimate bias can correlate with irrelevant nuisance variables. Concretely, we find the plug-in estimator bias tends to decrease with the effective dimensionality of neural populations. Thus if similarity between two populations appears to be explained by some manipulation of interest (e.g., training regime) it can be confounded through variation in effective dimensionality.

\begin{figure}[htb]
\centering
\includegraphics[width=.5\linewidth]{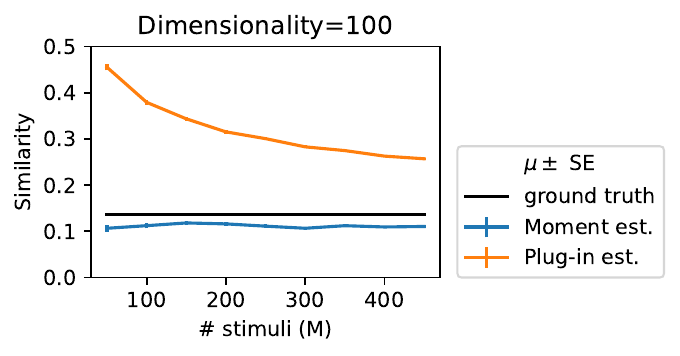}
\vspace{-0.25cm}
\caption{Comparison of plug-in and moment based estimators bias estimating similarity in the hidden layer of a deep neural network as a function of number of samples.  Here we specifically study a subset of the neurons ($N = 100$) in the penultimate layer of two ResNet-50 architectures trained on ImageNet with different initializations and training procedures.}
\label{fig:appendix-artificial-deep-net}
\end{figure}

A common question in the study  of neural network representations is how two networks with the same architecture trained on the same task but with different initializations and training procedures are similar/different from each other. 
To show how the estimators considered in this paper can be used as a tool to study this question, we considered two ResNet-50 \citep{he2016deep} architectures trained to categorize ImageNet \citep{deng2009imagenet}, specifically two sets of pretrained weights available in Pytorch (\texttt{ResNet50\_Weights.IMAGENET1K\_V1} and \texttt{ResNet50\_Weights.IMAGENET1K\_V2} as described \href{https://pytorch.org/vision/stable/models.html}{here}). 
We then compared a randomly chosen subset of the neurons (100 in each network) in the penultimate layer (before the final fully connected layer mapping to the logits) across the two networks using the plug-in estimator and the moment estimator. 
To compute a ground truth similarity metric we applied the plug-in estimator to the responses of these units across 432,064 images randomly chosen as a subset of the ImageNet dataset. 
To compare finite sample bias for each number of observed stimuli $M$,  we randomly re-sampled across images and calculated the mean and SE of the two estimators as a function of number of images (Fig.~\ref{fig:appendix-artificial-deep-net}). We found that the bias of the plug-in-estimator was at worst 3-fold and this bias decreased slowly, whereas the moment estimator showed a small amount of bias even with the smallest numbers of samples.

Finally, we considered how the bias of the plug-in estimator would vary with respect to irrelevant properties of the neural populations chosen. We reasoned that such a dependence  could confound results on similarity between neural populations. It is known that the effective dimensionality, ${(\sum_{i=1}^N \lambda_i)^2} / {\sum_{i=1}^N \lambda_i^2}$, of a response distribution determines the rate at which its sample covariance and thus singular values can be estimated. To determine if this in turn biased the plug-in estimator in a real application we randomly re-sampled with out replacement 100 units of the 2048 from the two neural networks 1000 times. We measured the ground truth similarity for each subset, the geometric mean of the effective dimensionality of the 100 units from the two networks (calculated across all images), and the plug-in average estimate across 50 random samplings of images. We found that the bias (difference of average plug-in estimate and ground truth), across re-sampling of units had a moderate negative correlation (r=-0.31) with the effective dimensionality of those populations. Thus observed differences in the similarity of neural network units may be confounded by their dimensionality and its effects on the plug-in estimator. 

\begin{figure}[htb]
\centering
\includegraphics[width=.4
\linewidth]{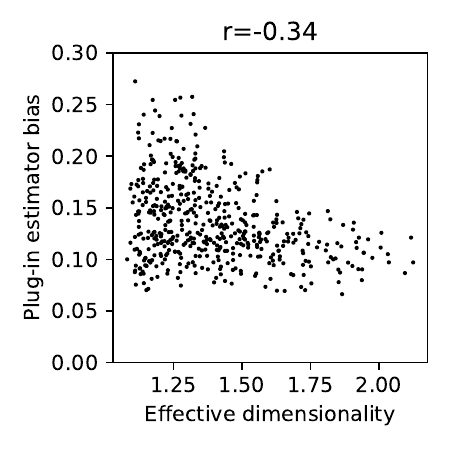}
\vspace{-0.25cm}
\caption{Each dot corresponds to a random selection of 100 neurons from the 2048 neurons in the penultimate layer of a ResNet-50. Across subsets of $N = 100$ neurons, we observe a negative correlation between the bias of the plug-in estimator with $M = 50$ stimuli and the effective dimensionality of the neural activations.}
\label{fig:appendix-dimensionality}
\end{figure}


\end{document}